\documentclass[a4paper]{article}
\usepackage{subcaption}
\usepackage{amssymb}
\usepackage{amsmath}
\usepackage[hidelinks]{hyperref}
\usepackage{graphicx}
\usepackage{fullpage}
\usepackage{tikz}
  \usetikzlibrary{shapes.geometric, calc, decorations.pathreplacing, positioning}
  \tikzstyle{tnnode}=[circle,
                      draw,
                      fill=teal!70,
                      text=white,
                      minimum size=0.5cm
                      ]
  \tikzstyle{isometry}=[isosceles triangle,
                      draw,
                      isosceles triangle apex angle=60,
                      shape border rotate=90,
                      fill=orange!60,
                      text=white,
                      minimum height=0.2cm,
                      minimum width=0.9cm,
                      inner sep=0pt
                      ]
  \tikzstyle{triangle}=[regular polygon, regular polygon sides=3, draw,
                        fill=teal!70, text=white, inner sep=2pt,
                        minimum size=0.78cm, shape border rotate=-90]
  \tikzstyle{bond}=[thick]
  \tikzstyle{phys}=[thick]
\usepackage{natbib}
  
\usepackage{fontawesome}

\newcommand{\footremember}[2]{%
    \footnote{#2}
    \newcounter{#1}
    \setcounter{#1}{\value{footnote}}%
}
\newcommand{\footrecall}[1]{%
    \footnotemark[\value{#1}]%
}

\newcommand{\nocontentsline}[3]{}
\newcommand{\tocless}[3]{\bgroup\let\addcontentsline=\nocontentsline#1{#2\label{#3}}\egroup}

\begin{document}

\title{Quantum-inspired tensor networks in machine learning models}

\author{
Guillermo Valverde\footnote{\scriptsize Corresponding author (\texttt{gvalverde\faAt{}vicomtech.org})}\,\,\,\footremember{affil1}{\scriptsize Vicomtech Foundation, Basque Research and Technology Alliance (BRTA), Donostia-San Sebastián 20009, Spain.}\,\footremember{affil2}{\scriptsize University of Deusto, Avda. de las Universidades, 24, Bilbao 48007, Spain.} 
\and
Igor García Olaizola\footrecall{affil1}\,\,\,\footrecall{affil2}
\and
Giannicola Scarpa\footremember{affil3}{\scriptsize Escuela T\'ecnica Superior de Ingenier\'ia de Sistemas Inform\'aticos, Universidad Polit\'ecnica de Madrid, Madrid 28031, Spain.}
\and 
Alejandro Pozas-Kerstjens\footremember{affil4}{\scriptsize Department of Applied Physics, University of Geneva, 1211 Geneva, Switzerland.}
}

\date{}

\maketitle

\begin{abstract}
    Tensor networks were developed in the context of many-body physics as compressed representations of multiparticle quantum states.
    These representations mitigate the exponential complexity of many-body systems by capturing only the most relevant dependencies.
    Due to the formal similarity between quantum entanglement and statistical correlations, tensor networks have recently been integrated in machine learning, operating both as alternative learning architectures and as decompositions of components of neural networks.
    The expectation is that the theoretical understanding of tensor networks developed within quantum many-body physics leads to novel methods that offer advantages in terms of computational efficiency, explainability, or privacy.
    Here we review the use of tensor networks in the context of machine learning, providing a critical assessment of the state of the art, the potential advantages, and the challenges that must be overcome.
\end{abstract}

\tableofcontents

\section{Introduction}\label{sec:introduction}
As a core branch of artificial intelligence, machine learning (ML) develops data-adaptive computational frameworks based on progressive refinement through experience.
In recent years, ML has yielded performance levels beyond human capability in domains such as visual computing and natural language processing \citep{krizhevsky2012,simonyan2014,vaswani2017}.
Among all ML models, deep neural networks (NNs) are the primary responsible for these developments.
Their modeling capacity derives from considerably over-parameterized architectures trained on large-scale datasets.
However, and despite huge practical success, neural networks also have limitations.
The most well-known one is that they rely on high computational and energy resources.
In addition to that, they are also predominantly opaque in their internal decision-making processes (black box) and can suffer from data leakage of sensitive information about their training through memorization and extraction attacks \citep{strubell2019energy,carlini2023extracting}.
These challenges motivate a natural question: can alternative learning paradigms solve these problems?

In this context, Tensor Networks (TNs) emerge as an alternative approach to address these challenges from a structural perspective.
Initially introduced in condensed matter physics as efficient representations of low-energy many-body quantum states \citep{white1992density}, TNs were designed to efficiently simulate quantum states and control the exponential growth of Hilbert spaces by factoring high-dimensional objects into low-rank tensor approximations \citep{verstraete2008matrix,orus2014practical,orus2019tensor}.
Transferring these tools to ML leads to a compelling analogy: quantum entanglement naturally applies to statistical correlations in classical data.
This makes TNs suitable for representing complex dependencies while explicitly controlling the complexity of the model through the dimension of the tensors \citep{Stoudenmire2016,Levine2019}.

Over the past decade, these techniques, typically restricted to problems in fundamental physics and quantum chemistry, have been successfully implemented in machine learning under the name of quantum-inspired machine learning.
In this context, there exist two main approaches.
In the first, TNs are used directly as learning architectures, where models such as Matrix Product States (MPS), Tree Tensor Networks (TTNs), or Projected Entangled Pair States (PEPSs) replace traditional ML models to perform tasks such as classification and generative modeling \citep{Stoudenmire2016, Stoudenmire2018, Chen2017}.
There is an optimistic view on these models, known as Tensor Neural Networks (TNNs), that is motivated by the expectation that the knowledge of TNs in the context of quantum many-body physics can be translated into the realm of ML, obtaining advantages in aspects such as privacy \citep{Pozas-Kerstjens2024A} or interpretability \citep{yao2021,Ran2023,aizpurua2025tensor}.
In the second, TNs are used as structured compression strategies for conventional NNs, where dense layers of pre-existing model weights are compressed to reduce the number of parameters and computational costs, without impacting the model performance in a significant manner \citep{novikov2015tensorizing,Tjandra2017,Gao2020}.
Recent studies have indicated that both approaches have found successful applications in numerous areas, such as deep learning \citep{Levine2019}, generative modeling \citep{Cheng2019}, computer vision \citep{Guala2023}, natural language processing \citep{Ma2019}, or the simulation of quantum computers \citep{Rudolph_2023}.

Beyond the compression properties that leads to both memory and computational efficiency, are there any fundamental differences between TNNs and conventional NNs? Indeed, TNs have two key properties that are particularly useful in a ML context.
First, TNs can be interpreted as explicit quantum states, granting direct access to quantum information theory quantities such as entanglement entropy and quantum mutual information.
These quantities provide intrinsic measures of the complexity and correlation of features within the model that enable interpretations that go beyond the post-hoc explanation techniques commonly used in ML \citep{Tangpanitanon2022,Ran2023}.
This perspective has been successfully explored in applications such as explainable natural language processing and cybersecurity, where TN observables offer quantitative information about what the model has learned \citep{aizpurua2025tensor}.
Second, tensor networks exhibit local transformations that keep the represented function invariant.
This property is typically an issue in conventional NNs, since they lead to trajectories in the parameter space that do not improve the loss while training.
However, in the context of TNNs, recent work has shown that these symmetries can be leveraged to reduce parameter-side information leakage by randomizing redundant degrees of freedom without degrading predictive performance \citep{Pozas-Kerstjens2024A,pareja2025}.
Unlike noise-based privacy mechanisms, this protection arises directly from the structure of the model.

The enthusiasm on the prospects of machine learning based on tensor networks has led to a large body of literature on quality applications.
However, this body of literature remains fragmented between communities and application domains.\footnote{One of the reasons for this, beyond the natural interest that the field raises, has roots in its very origin: TNs were discovered and developed in parallel and independently within the communities of quantum information \citep{fannes1992finitely,white1992density,vidal2003efficient} and of numerical analysis \citep{Oseledets2011I,Oseledets2011b}.}
The aim of this work is to analyze the current state of the art in the different uses of TNs in ML, highlighting successful applications, cases where technology-based approaches are insufficient, and gaps in the field.
Several reviews have already examined the broader area of tensor methods and TNs in ML under different lenses.
On the tensor decomposition side, \cite{Cichocki2016} and \cite{Cichocki2017} provide a comprehensive monograph on mathematical foundations and applications in optimization, \citet{Papalexakis_2016} reviews tensor methods for data mining and data fusion focusing on scalability, and \citet{ji2019survey} offers a more introductory overview of decompositions and their applications in ML.
On the TN side, \citet{Sengupta2022} gives a broad introduction to TN-based ML, \citet{Wang2023M} surveys the integration of TNs with NN architectures, and \cite{Ran2023} and \cite{Rieser2023} focus on quantum-inspired ML and quantum ML, respectively.
By contrast, our goal is to provide a critical assessment across application domains: we distinguish between native TN models and tensorized neural networks, discuss practical limitations, and ask where current claims are supported primarily by small-scale evidence.
We also survey the main software libraries and frameworks that make TN-based machine learning possible.
As part of this effort, we develop an interactive bibliographic map, which organizes the most influential articles in the field by author and topic, allowing the reader to easily navigate the literature \citep{valverde2026researchgraph}.

\section{Theoretical concepts} \label{sec:theory}
In this section we introduce the notation and several architectures useful for understanding the recent, tensor-based ML models that are appearing in the literature.
We assume familiarity with the algebraic concept of a tensor and the basic operations such as multiplication, transposition and contraction.
We refer the reader to \citet{biamonte2017tensornetworksnutshell} and \cite{Bridgeman2017} for more comprehensive introductions to the concepts discussed here.
Throughout this paper, we use the word \textit{order} to denote the number of indices (or legs) of a tensor, \text{dimension} for the size of an individual index, and \textit{shape} for the full collection of index dimensions.

\subsection{Penrose notation}
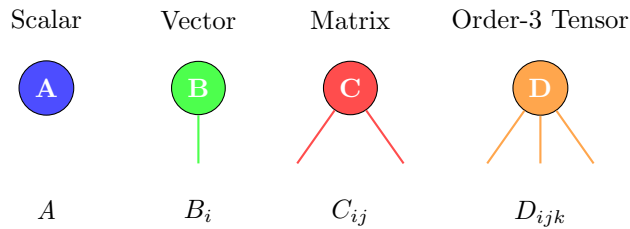
\begin{figure}[h]
    \centering
    \begin{tikzpicture}
        \node[draw, circle, fill=blue!70, text=white] (A) at (0, 0) {\textbf{A}};
        \node[below=1cm of A] {\textit{A}};
        \node[above=0.3cm of A] {Scalar};
    
        \node[draw, circle, fill=green!70, text=white] (B) at (2, 0) {\textbf{B}};
        \draw[thick, green!70] (B) -- ++(0, -1);
        \node[below=1cm of B] {\textit{$B_i$}};
        \node[above=0.3cm of B] {Vector};
    
        \node[draw, circle, fill=red!70, text=white] (C) at (4, 0) {\textbf{C}};
        \draw[thick, red!70] (C) -- ++(-0.7, -1);
        \draw[thick, red!70] (C) -- ++(0.7, -1);
        \node[below=1cm of C] {\textit{$C_{ij}$}};
        \node[above=0.3cm of C] {Matrix};
    
        \node[draw, circle, fill=orange!70, text=white] (D) at (6.5, 0) {\textbf{D}};
        \draw[thick, orange!70] (D) -- ++(0, -1);
        \draw[thick, orange!70] (D) -- ++(-0.7, -1);
        \draw[thick, orange!70] (D) -- ++(0.7, -1);
        \node[below=1cm of D] {\textit{$D_{ijk}$}};
        \node[above=0.3cm of D] {Order-3 Tensor};
    \end{tikzpicture}
    \caption{Penrose notation for a scalar, vector, matrix, and a order-3 tensor.}
    \label{fig:penrose1}
\end{figure}

Penrose notation \citep{Penrose1971R} emerged as a graphical language that simplifies the algebraic manipulations associated with tensors using visual diagrams of the operations and the relationships between them.
In this notation, each tensor is represented by a node (typically depicted by a geometric figure), with legs that represent the indices of the tensor.
Thus, a node without legs represents a scalar, a node with one leg represents a vector, a node with two legs represents a matrix, and so on (see Fig.~\ref{fig:penrose1}).
If legs from two nodes are connected, this indicates a contraction (i.e., a scalar product) between indices.

\begin{figure}[h]
    \centering
    \begin{tikzpicture}[
            grafo/.style={circle, draw, fill=teal!70, text=white, minimum size=0.8cm},
        every path/.style={thick}
    ]
        \node[draw, rectangle, minimum size=1cm, fill=teal!70, text=white] (T) at (-3,0) {$T$};
        
        \node at (-1.5,0) {\Huge $=$};
        
        \node[grafo] (A) at (0,0) {$A$};
        \node[grafo] (B) at (1,1) {$B$};
        \node[grafo] (C) at (2,0) {$C$};
        \node[grafo] (D) at (1,-1) {$D$};

        \draw (A) -- (B) node [below left = 6pt and 11pt] {$\alpha$};
        \draw (B) -- (C) node [above left = 7pt and 2pt] {$\beta$};
        \draw (C) -- (D) node [above right = 4pt and 11pt] {$\gamma$};
        \draw (D) -- (A) node [below right = 11pt and 4pt] {$\delta$};

        \draw (D) -- (1,-1.7) node [right] {$l$};
        \draw (A) -- (-0.7,0) node [above] {$j$};
        \draw (A) -- (-0.7,0.7) node [above] {$i$};
        \draw (A) -- (-0.7,-0.7) node [above] {$k$};
        \draw (T) -- (-4,0) node [above] {$j$}; 
        \draw (T) -- (-4,0.7) node [above] {$i$}; 
        \draw (T) -- (-4,-0.7) node [above] {$k$}; 
        \draw (T) -- (-3,-1) node [right] {$l$};
    \end{tikzpicture}
    \caption{Tensor network example. The matrices $B$ and $C$, the order-3 tensor $D$, and the order-5 tensor $A$ are multiplied along the relevant dimensions (i.e., contracted) to produce the order-4 tensor $T$.}
    \label{fig:complex_tensor}
\end{figure}
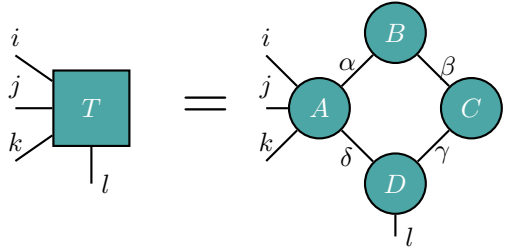

This graphical notation is very useful in simplifying the representation of calculations, since, when working with tensors, the equations can become complicated and long. Figure~\ref{fig:complex_tensor} shows a graphical representation of a complex TN composed of four tensors contracted together, which form a new, order-4 tensor.
This graphical representation corresponds to the equation $T_{ijkl}=A_{ijk\delta\alpha}B_{\alpha\beta}C_{\beta\gamma}D_{\delta l \gamma}$, where we follow Einstein's notation \citep{einstein1916} adding over repeated indices.

\subsection{Tensor operations}
\label{Sec:ten_op}
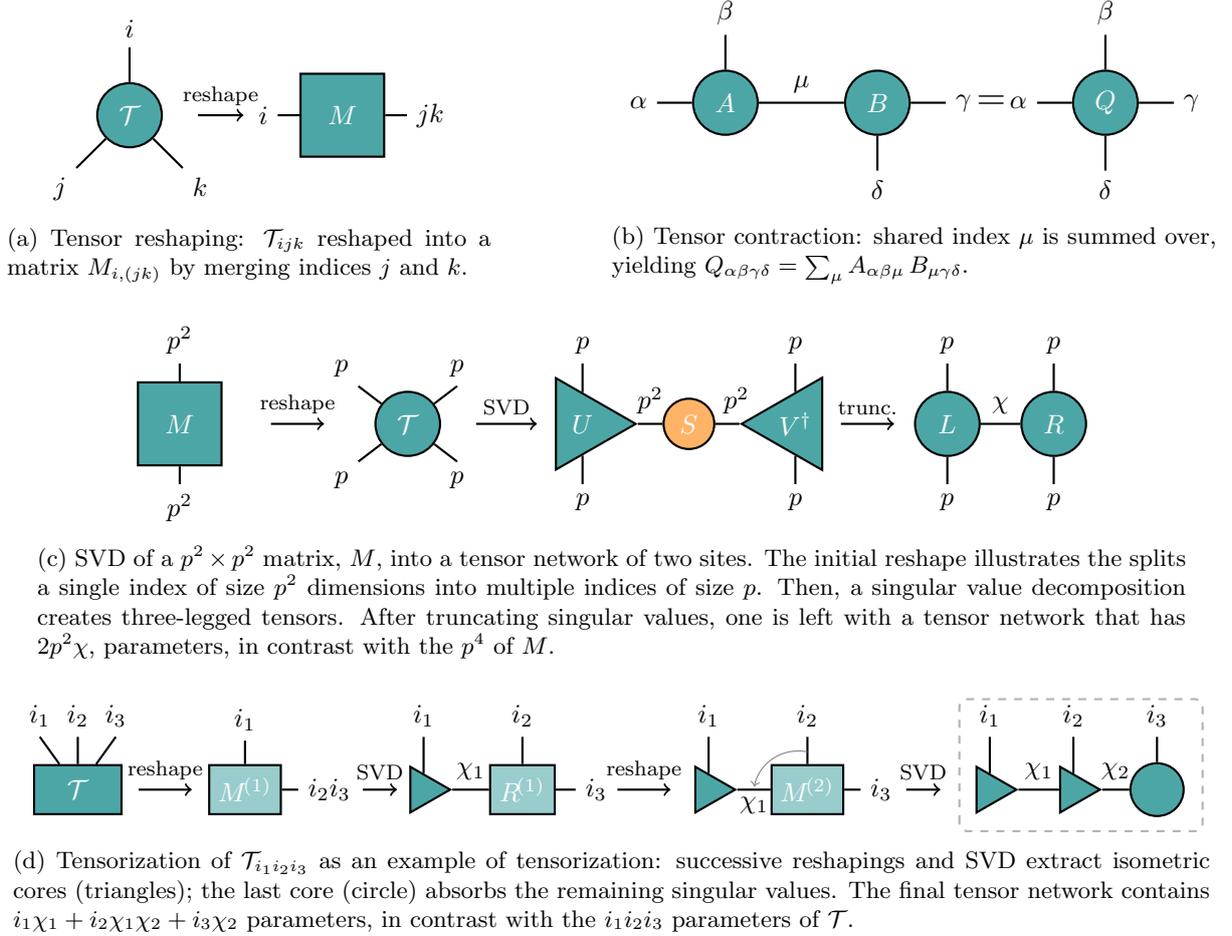
\begin{figure}[h]
    \centering
    \begin{subfigure}[b]{0.4\textwidth}
        \centering
        \begin{tikzpicture}[
            grafo/.style={circle, draw, fill=teal!70, text=white, minimum size=0.85cm},
            tensor/.style={draw, rectangle, minimum width=1.1cm, minimum height=1.1cm, 
                           fill=teal!70, text=white},
            every path/.style={thick}
        ]
            \node[grafo] (T) at (0,0) {$\mathcal{T}$};
            \draw (T) -- ++(0, 0.9)      node[above]       {$i$};
            \draw (T) -- ++(-0.7, -0.7)  node[below left]  {$j$};
            \draw (T) -- ++(0.7, -0.7)   node[below right]  {$k$};

            \draw[->, thick] (0.9, 0) -- (1.5, 0)
                node[midway, above] {\footnotesize reshape};

            \node[tensor] (M) at (2.8, 0) {$M$};
            \draw (M) -- ++(-0.85, 0) node[left]  {$i$};
            \draw (M) -- ++(0.85, 0)  node[right] {$jk$};
        \end{tikzpicture}
        \caption{Tensor reshaping: $\mathcal{T}_{ijk}$ reshaped into a matrix $M_{i,(jk)}$ by merging indices $j$ and $k$.}
        \label{fig:tensor_reshape}
    \end{subfigure}
    \hfill
    \begin{subfigure}[b]{0.5\textwidth}
        \centering
        \begin{tikzpicture}[
            grafo/.style={circle, draw, fill=teal!70, text=white, minimum size=0.85cm},
            every path/.style={thick}
        ]
            \node[grafo] (A) at (0,0) {$A$};
            \draw (A) -- ++(-0.9, 0) node[left] {$\alpha$};
            \draw (A) -- ++(0,  0.9) node[above] {$\beta$};

            \node[grafo] (B) at (2,0) {$B$};
            \draw (B) -- ++(0.9, 0) node[right] {$\gamma$};
            \draw (B) -- ++(0, -0.9) node[below] {$\delta$};
            \draw (A) -- (B) node[midway, above] {$\mu$};

            \node at (3.5, 0) {\Large $=$};

            \node[grafo] (Q) at (5,0) {$Q$};
            \draw (Q) -- ++(-0.9, 0) node[left] {$\alpha$};
            \draw (Q) -- ++(0,  0.9) node[above] {$\beta$};
            \draw (Q) -- ++(0.9, 0) node[right] {$\gamma$};
            \draw (Q) -- ++(0, -0.9) node[below] {$\delta$};
        \end{tikzpicture}
        \caption{Tensor contraction: shared index $\mu$ is summed over, yielding $Q_{\alpha\beta\gamma\delta} = \sum_{\mu} A_{\alpha\beta\mu}\, B_{\mu\gamma\delta}$.}
        \label{fig:tensor_contraction}
    \end{subfigure}

    \vspace{1.2em}

    \begin{subfigure}[b]{0.95\textwidth}
        \centering
        \begin{tikzpicture}[
            grafo/.style={circle, draw, fill=teal!70, text=white, minimum size=0.85cm},
            tensor/.style={draw, rectangle, minimum width=1.1cm, minimum height=1.1cm, fill=teal!70, text=white},
            isoU/.style={regular polygon, regular polygon sides=3, draw, fill=teal!70, text=white, inner sep=2pt, minimum size=1.4cm, shape border rotate=-90},
            isoV/.style={regular polygon, regular polygon sides=3, draw, fill=teal!70, text=white, inner sep=1pt, minimum size=0.85cm, shape border rotate=90},
            every path/.style={thick},
            nota/.style={<->, >=stealth, thin, gray, font=\scriptsize\itshape}
        ]
        \node[tensor] (M) at (-4.5, 0) {$M$};
        \draw (M) -- ++(0,  0.8) node[above] (p2a) {$p^2$};
        \draw (M) -- ++(0, -0.8) node[below] (p2b) {$p^2$};
    
        \draw[->, thick] (-3.3, 0) -- (-2.6, 0)
            node[midway, above] {\footnotesize reshape};

        \node[grafo] (T4) at (-1.5, 0) {$\mathcal{T}$};
        \draw (T4) -- ++(0.65,  0.5) node[above] (pa) {$p$};
        \draw (T4) -- ++(0.65, -0.5) node[below] {$p$};
        \draw (T4) -- ++(-0.65,  0.5) node[above left] (pb) {$p$};
        \draw (T4) -- ++(-0.65, -0.5) node[below left] {$p$};

        \draw[->, thick] (-0.6, 0) -- (0.2, 0)
            node[midway, above] {\footnotesize SVD};
    
        \node[isoU] (U) at (0.8, 0) {$U$};
        \draw (U) -- ++(0,  0.8) node[above] {$p$};
        \draw (U) -- ++(0, -0.8) node[below] {$p$};
    
        \node[circle, draw, fill=orange!60, text=white, minimum size=0.6cm] (S) at (2.2, 0) {$S$};
    
        \node[isoV] (V) at (3.6, 0) {$V^\dagger$};
        \draw (V) -- ++(0,  0.8) node[above] {$p$};
        \draw (V) -- ++(0, -0.8) node[below] {$p$};
    
        \draw (U.east) -- (S.west) node[midway, above] {$p^2$};
        \draw (S.east) -- (V.west) node[midway, above] {$\,\,\,\,p^2$};
    
        \draw[->, thick] (4.2, 0) -- (4.9, 0)
            node[midway, above] {\footnotesize trunc.};
    
        \node[grafo] (L) at (5.6, 0) {$L$};
        \draw (L) -- ++(0,  0.8) node[above] {$p$};
        \draw (L) -- ++(0, -0.8) node[below] {$p$};
    
        \node[grafo] (R) at (7, 0) {$R$};
        \draw (R) -- ++(0,  0.8) node[above] {$p$};
        \draw (R) -- ++(0, -0.8) node[below] {$p$};
    
        \draw (L) -- (R) node[midway, above] {$\chi$};
    \end{tikzpicture}
    \caption{SVD of a $p^2\times p^2$ matrix, $M$, into a tensor network of two sites. The initial reshape illustrates the splits a single index of size $p^2$ dimensions into multiple indices of size $p$. Then, a singular value decomposition creates three-legged tensors. After truncating singular values, one is left with a tensor network that has $2p^2\chi$, parameters, in contrast with the $p^4$ of $M$.}
    \label{fig:svd_mpo}
    \end{subfigure}

    \vspace{1.2em}

    \begin{subfigure}[b]{0.99\textwidth}
        \centering
        \begin{tikzpicture}[
            grafo/.style={circle, draw, fill=teal!70, text=white, minimum size=0.7cm},
            tensor/.style={draw, rectangle, minimum width=0.7cm, minimum height=0.65cm, fill=teal!70, text=white},
            remainder/.style={draw, rectangle, minimum width=0.7cm, minimum height=0.65cm, fill=teal!40, text=white},
            isoR/.style={regular polygon, regular polygon sides=3, draw, fill=teal!70, text=white, inner sep=1.5pt, minimum size=0.7cm, shape border rotate=-90},
            every path/.style={thick},
            sa/.style={->, thick, black},
            lbl/.style={},
        ]
        \node[tensor, minimum width=1.15cm] (T0) at (0, 0) {$\mathcal{T}$};
        \draw (T0) -- ++(-0.5, 0.7) node[above, lbl] {$i_1$};
        \draw (T0) -- ++(0, 0.7)    node[above, lbl] {$i_2$};
        \draw (T0) -- ++(0.5, 0.7)  node[above, lbl] {$i_3$};
    
        \draw[sa] (0.8, 0) -- (1.5, 0) node[midway, above] {\footnotesize reshape};
    
        \node[remainder, minimum width=0.9cm] (M1) at (2.2, 0) {$M^{\!(1)}$};
        \draw (M1) -- ++(0, 0.65) node[above, lbl] {$i_1$};
        \draw (M1) -- ++(0.7, 0)  node[right, lbl] {$i_2 i_3$};
    
        \draw[sa] (3.75, 0) -- (4.2, 0) node[midway, above] {\footnotesize SVD};
    
        \node[isoR] (A1) at (4.55, 0) {};
        \draw (A1) -- ++(0, 0.7) node[above, lbl] {$i_1$};
    
        \node[remainder, minimum width=0.85cm] (R1) at (5.85, 0) {$R^{\!(1)}$};
        \draw (R1) -- ++(0, 0.7) node[above, lbl] {$i_2$};
        \draw (R1) -- ++(0.7, 0)  node[right, lbl] {$i_3$};
        \draw (A1) -- (R1) node[midway, above, lbl] {$\chi_1$};
    
        \draw[sa] (7.1, 0) -- (7.8, 0) node[midway, above] {\footnotesize reshape};
    
        \node[isoR] (A1b) at (8.3, 0) {};
        \draw (A1b) -- ++(0, 0.7) node[above, lbl] {$i_1$};
    
        \node[remainder, minimum width=0.85cm] (M2) at (9.6, 0) {$M^{\!(2)}$};
        \draw (M2) -- (A1b); 
        
        \node (idx2) at (9.6, 0.7) [above, lbl] {$i_2$};
        \draw (M2) -- (idx2);
        \node (chi1_low) at (8.9, -0.2) [lbl] {$\chi_1$};
        
        \draw[<-, bend left=45, gray, thin] (chi1_low.north) to (9.6, 0.5);
    
        \draw (M2) -- ++(0.7, 0)  node[right, lbl] {$i_3$};
    
        \draw[sa] (10.9, 0) -- (11.35, 0) node[midway, above] {\footnotesize SVD};
    
        \node[isoR] (F1) at (12, 0) {};
        \draw (F1) -- ++(0, 0.7) node[above, lbl] {$i_1$};
    
        \node[isoR] (F2) at (13.1, 0) {};
        \draw (F2) -- ++(0, 0.7) node[above, lbl] {$i_2$};
    
        \node[grafo] (F3) at (14.2, 0) {};
        \draw (F3) -- ++(0, 0.7) node[above, lbl] {$i_3$};
    
        \draw (F1) -- (F2) node[midway, above, lbl] {$\chi_1$};
        \draw (F2) -- (F3) node[midway, above, lbl] {$\chi_2$};
    
        \draw[dashed, rounded corners=3pt, gray!60]
            (11.6, -0.55) rectangle (14.8, 1.2);
    
        \end{tikzpicture}
        \caption{Tensorization of $\mathcal{T}_{i_1 i_2 i_3}$ as an example of tensorization: successive reshapings and SVD extract isometric cores (triangles); the last core (circle) absorbs the remaining singular values. The final tensor network contains $i_1\chi_1+i_2\chi_1\chi_2+i_3\chi_2$ parameters, in contrast with the $i_1i_2i_3$ parameters of $\mathcal{T}$.}
        \label{fig:mps_decomposition}
    \end{subfigure}
    \caption{Fundamental tensor network operations: (a)~reshaping, (b)~tensor contraction, (c)~SVD decomposition of a high-dimensional matrix, and (d)~tensorization into a linear tensor network.}
    \label{fig:tensor_operations}
\end{figure}

The first operation to consider is \textit{tensor reshaping}, which consists of reorganizing the indices of a tensor without altering its values.
By modifying the grouping of indices, reshaping allows the tensor to be transformed into the most suitable format for each use case.
This operation is particularly useful because it allows transforming arbitrary tensors into matrices (see Fig.~\ref{fig:tensor_reshape}), to which matrix decompositions, such as singular value decomposition (SVD, illustrated in Fig.~\ref{fig:svd_mpo}) or the QR decomposition, can be applied afterwards.
This enables the efficient analysis and manipulation of tensor structures in many applications.

A strategic application of this process is \textit{tensorization}, by which arbitrary tensors are given a TN description or approximation.
Tensorization is shown in Figs.~\ref{fig:svd_mpo} and \ref{fig:mps_decomposition}.
One begins by reshaping the indices of the tensor into a matrix, and applies matrix decomposition algorithms to isolate one of the indices.
Then, the resulting matrices follow the same procedure, continuing iteratively until the dimensions are small enough.
Let us illustrate this process with a vector in a high-dimensional space (say, $d^n$).
As a first step, we can reshape this vector into a matrix of shape $d\times d^{n-1}$.
Applying SVD to this matrix, we are left with a small $d\times d$ matrix and a larger one, of shape $d\times d^{n-1}$.
For the latter, we can now reshape it as a $d^2\times d^{n-2}$ matrix, in which SVD produces a $d^2\times d^2$ matrix and a $d^2\times d^{n-2}$ matrix.
The smaller matrix can be reshaped into a three-legged tensor of shape $d\times d\times d^2$ that can contract with the small matrix of the previous step.
As for the larger one, one continues applying reshaping and SVD.
The final result is a linear tensor network (which we discuss in more depth in Section~\ref{sec:mps}).

It is not difficult to see that the process described above does not reduce the amount of memory needed to store the tensor, since the dimension of the internal legs of the tensor network increases exponentially with the number of steps.
In order to achieve savings, after the SVD one has to truncate the virtual or bond dimension.
This is typically done by applying a projection over the largest singular values.
By the Eckart-Young theorem \citep{eckart1936approximation}, this way of proceeding gives the best fixed-rank approximation in Fr\"obenius norm of every truncated matrix.
This results in an approximation error of the original vector that scales with the sum of the squares of the singular values discarded \citep{verstraete2006}, and thus it scales linearly with the number of sites in the resulting tensor network.
However, this local optimality does not, in general, guarantee a globally optimal low-rank approximation of the full tensor.
Iterative sweep strategies, in which successive SVD truncations are performed back and forth across the network, can improve the quality of the global approximation \citep{schollwock2011density}.
Low-rank approximations based on linear TNs can be obtained as well through variational methods, which are more computationally demanding but yield globally better approximations.
The most popular is the Density Matrix Renormalization Group (DMRG) algorithm \citep{white1992density,schollwock2011density}, which iteratively optimizes the TN cores by solving eigenvalue problems locally.
An alternative is Riemannian optimization \citep{uschmajew2013geometry}, which exploits the smooth manifold structure of TN architectures, avoiding rank growth while allowing flexible loss functions beyond energy minimization.

It is important to note that reshaping alters the index structure of the tensor.
If spatial or temporal structure was being stored in the index structure of the tensor, reshaping may destroy it.
This can be easily observed, for instance, when flattening an image to a vector: while pixels in adjacent columns keep being adjacent in the vector, this is not the case for pixels in adjacent rows.
Thus, the selection of the \textit{reshape} method directly affects the compression capacity of the model.
Instead, new approaches consider tensor algebra to operate on multidimensional data directly \citep{Kilmer2021}.
These methodologies guarantee optimal representation and compression of the information, allowing data manipulation while preserving all its geometric and temporal correlation without the need for reshaping.

Another relevant concept is the \textit{contraction} of an edge (see Fig.~\ref{fig:tensor_contraction}).
Contraction generalizes the matrix product by fusing two connected nodes into a single tensor that inherits all their unshared indices.
While this reduces the number of nodes in the tensor network, the resulting tensor can be much heavier than the original factors if the contracted index $n$ is small relative to the remaining free indices ($m, p$).
Therefore, contracting an edge simplifies the network topology but can significantly increase the memory consumption of the resulting node.
This is the operation that allows us to recover the dense tensor if we have the low-rank representation.
Often, the choice of the order of this contraction is critical for achieving computational cost savings.

However, when these architectures are used in deep learning, gradients need to be calculated to train the network.
This is a critical point in their adaptation to state of the art architectures, and it has a different treatment depending on the type of architecture.
Tensorized Deep Learning architectures can be directly trained with the standard Machine Learning frameworks, such as PyTorch \citep{ansel2024}, since the tensor contractions involved are reduced to sequences of matrix product, for which these frameworks provide efficient automatic differentiation.
By contrast, native TN architectures involve arbitrary contraction graphs.
Computing gradients through these contraction graphs requires differentiating with respect to the contraction order and structure itself, for which \citet{Liao_2019} introduced dedicated algorithms.

\subsection{Gauge freedom and canonical forms}
\label{sec:gauge}
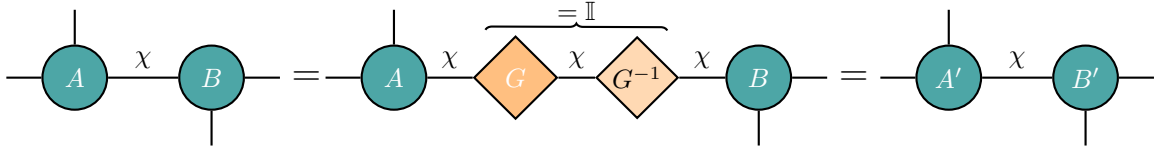
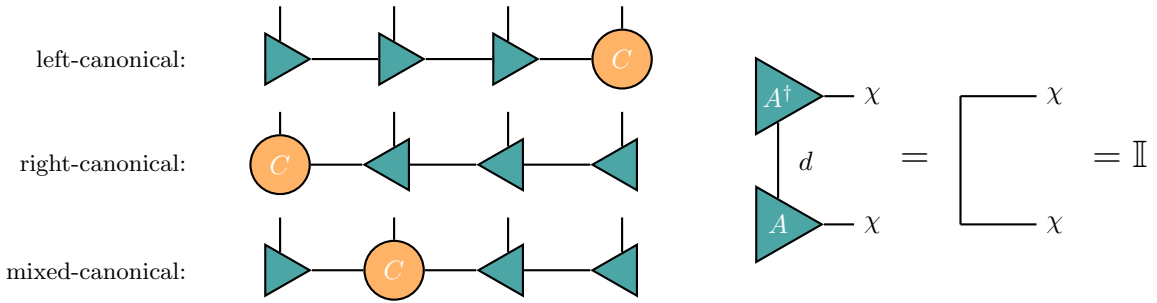
\begin{figure}[h]
    \centering
    \begin{subfigure}[b]{0.95\textwidth}
        \centering
        \begin{tikzpicture}[
            grafo/.style={circle, draw, fill=teal!70, text=white, minimum size=0.85cm},
            every path/.style={thick}
        ]
            \node[grafo] (A) at (0,0) {$A$};
            \node[grafo] (B) at (1.8,0) {$B$};
            \draw (A) -- ++(-0.9, 0);
            \draw (A) -- ++(0,  0.9);
            \draw (B) -- ++(0.9, 0);
            \draw (B) -- ++(0, -0.9);
            \draw (A) -- (B) node[midway, above] {$\chi$};

            \node at (3.05, 0) {\Large $=$};

            \node[grafo] (A2) at (4.2, 0) {$A$};
            \node[draw, diamond, fill=orange!50, text=white,
                  minimum size=1.1cm, inner sep=1pt] (G)  at (5.8, 0) {$G$};
            \node[draw, diamond, fill=orange!30, text=black,
                  minimum size=0.8cm, inner sep=1pt] (Gi) at (7.4, 0) {$G^{-1}$};
            \node[grafo] (B2) at (9., 0) {$B$};

            \draw (A2) -- ++(-0.9, 0);
            \draw (A2) -- ++(0,  0.9);
            \draw (B2) -- ++(0.9, 0);
            \draw (B2) -- ++(0, -0.9);
            \draw (A2) -- (G)  node[midway, above] {$\chi$};
            \draw (G)  -- (Gi) node[midway, above] {$\chi$};
            \node (braceleft) at ([xshift=-15,yshift=2]G.north) {};
            \node (braceright) at ([xshift=15,yshift=2]Gi.north) {};
            \draw[decoration={brace},decorate] (braceleft) -- (braceright) node[midway, above] {$=\mathbb{I}$};
            \draw (Gi) -- (B2) node[midway, above] {$\chi$};

            \node at (10.25, 0) {\Large $=$};

            \node[grafo] (A3) at (11.5, 0) {$A'$};
            \node[grafo] (B3) at (13.3, 0) {$B'$};
            \draw (A3) -- ++(-0.9, 0);
            \draw (A3) -- ++(0,  0.9);
            \draw (B3) -- ++(0.9, 0);
            \draw (B3) -- ++(0, -0.9);
            \draw (A3) -- (B3) node[midway, above] {$\chi$};

        \end{tikzpicture}
        \caption{Gauge freedom: inserting $G\,G^{-1}=\mathbb{I}$ on any virtual bond
                 leaves the global tensor unchanged while transforming local tensors
                 $A \to A'=AG$ and $B \to B'=G^{-1}B$.}
        \label{fig:gauge_freedom}
    \end{subfigure}

    \vspace{1.5em}

    \begin{subfigure}[b]{0.59\textwidth}
        \centering
        \begin{tikzpicture}[
            isoL/.style={regular polygon, regular polygon sides=3, draw,
                         fill=teal!70, text=white, inner sep=2pt,
                         minimum size=0.78cm, shape border rotate=-90},
            isoR/.style={regular polygon, regular polygon sides=3, draw,
                         fill=teal!70, text=white, inner sep=2pt,
                         minimum size=0.78cm, shape border rotate=90},
            center/.style={circle, draw, fill=orange!60, text=white,
                           minimum size=0.78cm},
            every path/.style={thick}
        ]

            \node[font=\small, anchor=east] at (-0.8, 2.8) {left-canonical:};

            \node[isoL]  (LL1) at (0.3, 2.8) {};
            \node[isoL]  (LL2) at (1.8, 2.8) {};
            \node[isoL]  (LL3) at (3.3, 2.8) {};
            \node[center](LL4) at (4.8, 2.8) {$C$};

            \draw (LL1) -- ++(0, 0.7);
            \draw (LL2) -- ++(0, 0.7);
            \draw (LL3) -- ++(0, 0.7);
            \draw (LL4) -- ++(0, 0.7);

            \draw (LL1) -- (LL2);
            \draw (LL2) -- (LL3);
            \draw (LL3) -- (LL4);

            \node[font=\small, anchor=east] at (-0.8, 1.4) {right-canonical:};

            \node[center](RR1) at (0.3, 1.4) {$C$};
            \node[isoR]  (RR2) at (1.8, 1.4) {};
            \node[isoR]  (RR3) at (3.3, 1.4) {};
            \node[isoR]  (RR4) at (4.8, 1.4) {};

            \draw (RR1) -- ++(0, 0.7);
            \draw (RR2) -- ++(0, 0.7);
            \draw (RR3) -- ++(0, 0.7);
            \draw (RR4) -- ++(0, 0.7);

            \draw (RR1) -- (RR2);
            \draw (RR2) -- (RR3);
            \draw (RR3) -- (RR4);

            \node[font=\small, anchor=east] at (-0.8, 0) {mixed-canonical:};

            \node[isoL]  (ML1) at (0.3, 0) {};
            \node[center](MC)  at (1.8, 0) {$C$};
            \node[isoR]  (MR1) at (3.3, 0) {};
            \node[isoR]  (MR2) at (4.8, 0) {};

            \draw (ML1) -- ++(0, 0.7);
            \draw (MC)  -- ++(0, 0.7);
            \draw (MR1) -- ++(0, 0.7);
            \draw (MR2) -- ++(0, 0.7);

            \draw (ML1) -- (MC);
            \draw (MC)  -- (MR1);
            \draw (MR1) -- (MR2);

        \end{tikzpicture}
        \caption{The three canonical forms of an MPS: left-canonical ($\triangleright$),
                 right-canonical ($\triangleleft$), and mixed-canonical with
                 orthogonality center $C$ (orange) at position 2.}
        \label{fig:canonical_forms}
    \end{subfigure}
    \hfill
    \begin{subfigure}[b]{0.36\textwidth}
        \centering
        \begin{tikzpicture}[
            isoL/.style={regular polygon, regular polygon sides=3, draw,
                         fill=teal!70, text=white, inner sep=2pt,
                         minimum size=0.78cm, shape border rotate=-90},
            every path/.style={thick}
        ]
            \node[regular polygon, regular polygon sides=3, draw,
                  fill=teal!70, text=white, inner sep=0.01pt,
                  minimum size=0.78cm, shape border rotate=-90] (Ad) at (0, 1.7) {$A^\dagger$};
            \node[regular polygon, regular polygon sides=3, draw,
                  fill=teal!70, text=white, inner sep=1pt,
                  minimum size=1.15cm, shape border rotate=-90] (Al) at (0, 0.0) {$A$};

            \draw (Ad) -- (Al) node[midway, right=4pt] {$d$};

            \draw (Ad) -- ++(1.0, 0) node[right] {$\chi$};
            \draw (Al) -- ++(1.0, 0) node[right] {$\chi$};

            \node at (1.8, 0.85) {\Large $=$};

            \draw[thick] (2.4, 0.0) -- (2.4, 1.7);
            \draw[thick] (2.4, 1.7) -- ++(1.0, 0) node[right] {$\chi$};
            \draw[thick] (2.4, 0.0) -- ++(1.0, 0) node[right] {$\chi$};

            \node at (4.5, 0.93) {\Large$=\mathbb{I}$};

        \end{tikzpicture}
        \vspace{.5cm}
        \caption{Isometry property: contracting $A$ with $A^\dagger$ over the physical
                 index $d$ yields the identity on the virtual space $\chi$.}
        \label{fig:isometry}
    \end{subfigure}

    \caption{Gauge freedom and canonical forms: (a) inserting $G\,G^{-1}$ on a
             virtual bond leaves the global tensor invariant; (b) the three canonical
             forms of an MPS; (c) isometry property $A^\dagger A = \mathbb{I}$.}
    \label{fig:gauge_canonical}
\end{figure}

One of the defining features of TNs, that distinguishes them from other ML architectures, is gauge freedom.
Since TNs are based on scalar products, it is always possible to insert between two nodes an identity matrix expressed as the product of any invertible matrix and its inverse \citep{biamonte2017tensornetworksnutshell}, as illustrated in Fig.~\ref{fig:gauge_freedom}.
This operation does not modify the global tensor, but it transforms the local tensors, enabling the same global tensor to be represented by an infinite number of equivalent sets of local tensors.

Rather than being a drawback, this redundancy becomes a powerful mathematical tool, since it allows us to enforce desirable properties on the tensors that simplify computations.
\textit{Canonical forms} arise in this context.
For example, in MPS (see Section~\ref{sec:mps}), it is possible to make all tensors isometric (i.e. $A^\dagger A = \mathbb{I}$, see Fig.~\ref{fig:isometry}) except one, which acts as the orthogonality center. 
In this case, the contraction of the entire network with itself can be reduced to the contraction of this center with itself \citep{orus2014practical}.
Notably, this form is easy to obtain through sequences of SVD (see Fig.~\ref{fig:canonical_forms}).
Tree structured TNs admit an analogous exact canonicalization, since the absence of loops allows SVD sweeps to be performed without ambiguity \citep{shi2006classical}. For more complex architectures with loops, such as PEPS (see section~\ref{sec:peps}), an exact orthogonality center cannot be defined in general, although canonical forms have recently been developed \cite{acuaviva2023}.
In these architectures, it is possible to define \textit{quasi-canonical forms} can be approximated via message passing algorithms such as Belief Propagation \citep{alkabetz2021tensor,tindall2023}.

One notable consequence of gauge freedom is its consequences for privacy and model obfuscation in ML.
Because gauge transformations leave the global tensor unchanged while modifying its internal representation, the parameters of the model can be rewritten in many equivalent ways.
This makes the internal structure harder to interpret or reverse-engineer, effectively hiding information encoded in the individual nodes without affecting the functionality of the model \citep{Pozas-Kerstjens2024A,pareja2025}.
In this sense, TNs naturally provide a form of structural protection against some forms of data extraction \citep{carlini2023extracting}, a property that will be discussed in more detail in the following sections.

\subsection{Tensor network architectures}
Any network of connected tensors is a TN.
However, one can identify in the literature several concrete families of TNs, whose structure is particularly suitable for performing calculations efficiently, capuring specific correlation patterns, or for concrete applications.
Matrix Product States (MPS) and Matrix Product Operators (MPO) are linear networks that are widely used due to their simplicity and computational efficiency \citep{Oseledets2011I,Cichocki2016}.
For systems with periodic conditions or cyclic patterns, Ring TNs (also known as Tensor Rings) modify the MPS by closing the chains \citep{Cheng2020,Liu_2020}.
For data with a two-dimensional structure, such as images, Projected Entangled Pair States (PEPS) extend the network by forming a grid \citep{Cheng2021}.
To model hierarchical or long-range interactions, Tree Tensor Networks (TTN) organize tensors into a branched tree structure \citep{Cheng2019,Reyes2021}, while the Multiscale Entanglement Renormalization Ansatz (MERA) refines this idea by adding disentanglement tensors to efficiently capture scale invariance properties and critical systems \citep{Kong2021,Hou2024}.
In this section, we discuss in more detail each of these architectures and describe their applications in the context of ML. 

\tocless\subsubsection{Matrix Product State / Tensor Train}
\begin{figure}[h]
    \centering
    \begin{tikzpicture}
        \node (A1) [tnnode] at (0,0) {};
        \node (A2) [tnnode] at (1,0) {};
        \node (A3) [tnnode] at (2,0) {};
        \node (A4) [tnnode] at (3,0) {};
        \node (A5) [tnnode] at (4,0) {};
        
        \draw[bond] (A1) -- (A2);
        \draw[bond] (A2) -- (A3);
        \draw[bond] (A3) -- (A4);
        \draw[bond] (A4) -- (A5);

        \draw[phys] (A1) -- ++(0,-1);
        \draw[phys] (A2) -- ++(0,-1);
        \draw[phys] (A3) -- ++(0,-1);
        \draw[phys] (A4) -- ++(0,-1);
        \draw[phys] (A5) -- ++(0,-1);
    \end{tikzpicture}
    \caption{A matrix product state of five sites.}
    \label{fig:mps_extended}
\end{figure}
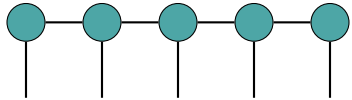

{sec:mps}
Matrix product states, also known as Tensor Trains (TT), are a class of TNs that decompose a tensor of $N$ indices into a chain of $N$ tensors of three indices, as in Fig.~\ref{fig:mps_extended}.
This structure was developed in parallel in the communities studying quantum many-body systems, where it has been instrumental in the development of efficient algorithms for the study of strongly correlated systems (see the review of \cite{cirac2021}), and of numerical analysis \citep{grasedyck2010,hackbusch2012}.

The free indices are called physical indices, and their dimension is the physical dimension, while the contracted indices of the MPS representation determine its bond dimension.
A tensor with \(N\) free indices, each of physical dimension \(d\), contains a total of \(d^N\) parameters.
However, when represented as an MPS with bond dimension \(\chi\), the number of parameters is reduced to approximately \(N d \chi^2\).

It is possible to give an MPS/TT representation of any tensor via a sequence of reshapings and singular value decompositions (see Fig.~\ref{fig:mps_decomposition}).
When cutting the singular values obtained this way to only $\chi$ of them, the $\ell_2$ error incurred corresponds to the sum of the singular values that have been left out \citep{verstraete2006}.
Thus, the fidelity of a low-rank approximation depends on the degree of correlation present in the data: the higher the correlation, the larger are the singular values that are discarded, and thus the greater the error of making a low-rank approximation.
In other words, the higher the correlation within the tensor, the larger the bond dimension required to capture its structure by a tensor network.

Another advantage of the MPS representation is the efficiency in computing the scalar product between tensors.
The direct contraction of two tensors of \(N\) free indices with physical dimension \(d\), presents a complexity of \(\mathcal{O}(d^N)\).
In contrast, when the tensors are represented in MPS format with bond dimension \(\chi\), the computational complexity is reduced to \(\mathcal{O}(N \chi^3 d)\), allowing for much more efficient processing on large systems.

The first proposals for using MPS in machine learning date back to 2016 \citep{Stoudenmire2016,novikov2018exponential} as model architectures for supervised learning tasks.
In particular, \cite{Stoudenmire2016} demonstrated that MPS can be efficiently optimized, using physics-inspired strategies.
Since then, they have also been used, e.g., in unsupervised learning to model distributions \citep{Liu_2023} and for hyperspectral-multiespectral image fusion \citep{Yang2023}.
Other relevant applications of these architectures include improved interpretability \citep{aizpurua2025tensor} and enhanced privacy \citep{Pozas-Kerstjens2024A}.

\tocless\subsubsection{Matrix Product Operator}
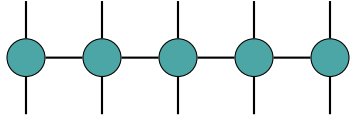
\begin{figure}[h]
\centering
    \begin{tikzpicture}
        \node (A1) [tnnode] at (-2,0) {};
        \node (A2) [tnnode] at (-1,0) {};
        \node (A3) [tnnode] at (0,0) {};
        \node (A4) [tnnode] at (1,0) {};
        \node (A5) [tnnode] at (2,0) {};

        \draw[bond] (A1) -- (A2);
        \draw[bond] (A2) -- (A3);
        \draw[bond] (A3) -- (A4);
        \draw[bond] (A4) -- (A5);

        \draw[phys] (A1) -- ++(0,0.75);
        \draw[phys] (A2) -- ++(0,0.75);
        \draw[phys] (A3) -- ++(0,0.75);
        \draw[phys] (A4) -- ++(0,0.75);
        \draw[phys] (A5) -- ++(0,0.75);
        \draw[phys] (A1) -- ++(0,-0.75);
        \draw[phys] (A2) -- ++(0,-0.75);
        \draw[phys] (A3) -- ++(0,-0.75);
        \draw[phys] (A4) -- ++(0,-0.75);
        \draw[phys] (A5) -- ++(0,-0.75);

    \end{tikzpicture}
    \caption{A matrix product operator of five sites.}
    \label{fig:mpo_extended}
\end{figure}

{}
Matrix product operators are an extension of MPS that allows the representation of linear operators, such as Hamiltonians or density matrices in the context of quantum mechanics, within the framework of TNs.
If MPS can be understood as efficient approximations of high-dimensional vectors (recall Sec.~\ref{Sec:ten_op}), then MPO constitute their natural extension for the representation of matrices in that same space.
Thus, while an MPS has \(N\) indices associated with the degrees of freedom of the system, an MPO has \(N\) input indices and \(N\) output indices, as shown in Fig.~\ref{fig:mpo_extended}.

The number of parameters in an MPO representation scales like \(\mathcal{O}(N d^2 \chi^2)\).
In terms of computational complexity, applying an MPO to an MPS scales as \(\mathcal{O}(N \chi ^6 d)\).
In contrast, a na\"ive dense representation would treat the problem as a vector–matrix multiplication, leading to a scaling of \(\mathcal{O}(d^{2N})\).

Among the most common applications of MPO is the tensorization of neural network layers \citep{novikov2015tensorizing,Gao2020}.
This is, the linear layers in deep neural networks are given an MPO representation and the matrix-vector products in the forward pass are substituted by MPO-MPS contractions.
This leads to a substantial reduction in the number of parameters describing a model, as well as an improvement in computational efficiency during inference.
This approach has been applied to, e.g., video processing and analysis models \citep{Yang2017,Zhao2020} and recurrent neural networks (RNNs) \citep{Tjandra2017}.

\tocless\subsubsection{Projected Entangled Pair States}{sec:peps}
Projected Entangled Pair States are widely regarded as the natural two-dimensional extension of MPS \citep{verstraete2008matrix}.
They are tensor networks defined on two-dimensional lattices, where each lattice site is associated with a local tensor comprising one physical index that corresponds to the data dimension, and with several virtual indices that connect to neighboring sites.
Due to this intrinsic two-dimensional connectivity, PEPS are particularly well suited for representing states and data exhibiting spatial locality and area-law correlation structures, which are characteristic of many two-dimensional quantum systems \citep{verstraete2008matrix,orus2014practical}.

\begin{figure}[h]
    \centering
    \begin{tikzpicture}
        \def\ax{1.8}  \def\ay{0.9}   
        \def\bx{-1.8} \def\by{0.9}   
        
        \foreach \i in {0,1,2}{
          \foreach \j in {0,1,2}{
            \coordinate (p-\i-\j) at ({\i*\ax+\j*\bx},{\i*\ay+\j*\by});
            \node[tnnode] (A-\i-\j) at (p-\i-\j) {};
            \draw[phys] (A-\i-\j) -- ++(0,0.75);
          }
        }

        \foreach \i in {0,1}{
          \foreach \j in {0,1,2}{
            \draw[bond] (A-\i-\j) -- (A-\the\numexpr\i+1\relax-\j);
          }
        }
        \foreach \i in {0,1,2}{
          \foreach \j in {0,1}{
            \draw[bond] (A-\i-\j) -- (A-\i-\the\numexpr\j+1\relax);
          }
        }
    \end{tikzpicture}
    \caption{PEPS lattice of size three.}
    \label{fig:peps_3x3}
\end{figure}
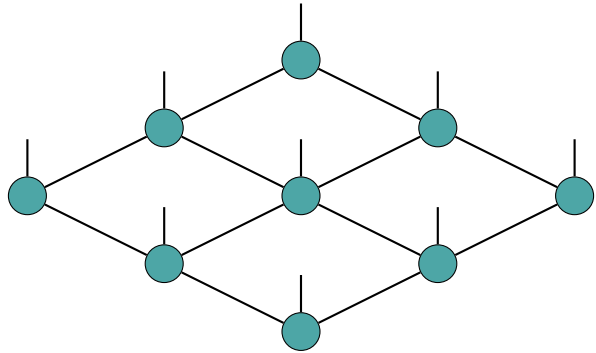

Unlike MPS, however, the exact contraction of general PEPS is computationally demanding \citep{schuch2007,haferkamp2020,Scarpa2020}, and their practical application, therefore, relies on approximate contraction schemes or renormalization-inspired methods \citep{cirac2021}.

In the context of machine learning, PEPS have been explored as supervised models for image classification, since in this case the model can faithfully capture the 2-D correlations that are present in the data.
Indeed, models based on PEPS have demonstrated performance comparable to that of traditional architectures while using fewer parameters and have shown advantages over tree-structured tensor networks on standard benchmarks such as MNIST and Fashion-MNIST \citep{Cheng2021}.

\tocless\subsubsection{Tree Tensor Networks}{sec:ttn}
Tree tensor networks, as their name suggests, are tensor networks defined by a hierarchical, tree-like structure.
In a TTN, nodes are arranged across successive layers without forming closed loops.
Unlike PEPS, this design allows for efficient contraction, which makes TTNs computationally appealing for large-scale applications.
Because information is aggregated locally at each hierarchical layer, TTNs intrinsically realize a multiscale representation of data, thereby enabling the model to extract salient features across progressively distinct levels of resolution \citep{Stoudenmire2018,Liu2019}.

\begin{figure}[h]
    \centering
    \begin{tikzpicture}    

        \node[tnnode] (L1) at (0,0) {};
        \node[tnnode] (L2) at (1,0) {};
        \node[tnnode] (L3) at (2,0) {};
        \node[tnnode] (L4) at (3,0) {};
        \node[tnnode] (L5) at (4,0) {};
        \node[tnnode] (L6) at (5,0) {};
        \node[tnnode] (L7) at (6,0) {};
        \node[tnnode] (L8) at (7,0) {};

        \node[isometry] (M1) at (0.5, 1.) {};
        \node[isometry] (M2) at (2.5, 1.) {};
        \node[isometry] (M3) at (4.5, 1.) {};
        \node[isometry] (M4) at (6.5, 1.) {};
        
        \draw[bond] (L1) -- (M1.left corner);
        \draw[bond] (L2) -- (M1.right corner);
        
        \draw[bond] (L3) -- (M2.left corner);
        \draw[bond] (L4) -- (M2.right corner);
        
        \draw[bond] (L5) -- (M3.left corner);
        \draw[bond] (L6) -- (M3.right corner);
        
        \draw[bond] (L7) -- (M4.left corner);
        \draw[bond] (L8) -- (M4.right corner);

        \node[isometry] (U1) at (1.5, 2.) {};
        \node[isometry] (U2) at (5.5, 2.) {};
        
        \draw[bond] (M1.apex) -- (U1.left corner);
        \draw[bond] (M2.apex) -- (U1.right corner);
        
        \draw[bond] (M3.apex) -- (U2.left corner);
        \draw[bond] (M4.apex) -- (U2.right corner);

        \node[isometry] (R) at (3.5, 3.) {};
        
        \draw[bond] (U1.apex) -- (R.left corner);
        \draw[bond] (U2.apex) -- (R.right corner);

        \draw[phys] (R.apex) -- ++(0, 0.5);
        
    \end{tikzpicture}
    \caption{Tree tensor network with eight leaf nodes.}
    \label{fig:ttn_8}
\end{figure}
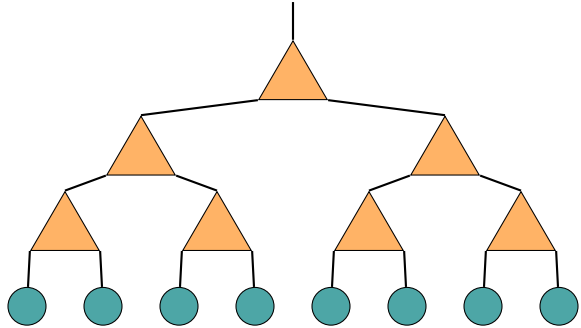

In ML applications, TTNs have found utility in both generative and discriminative settings.
Their effectiveness in generative modeling is largely attributable to the tractable probabilistic structure inherent to the tree-based architecture \citep{Cheng2019}.
In supervised learning scenarios, TTN-based classifiers have demonstrated competitive predictive performance together with substantial stability \citep{Wall2021}.
More recent investigations have tried to enhance robustness and refine regularization through techniques such as rank constraints and tensor dropout \citep{Chen_2024}, while concurrently examining how noise and decoherence affect tensor-network-based learning models \citep{Liao2023}.
Although TTNs are inherently less expressive than TNs with loops in capturing long-range correlations, they nevertheless provide a principled and scalable inductive bias that is well suited to multiscale learning problems.

\tocless\subsubsection{Multiscale Entanglement Renormalization Ansatz}{sec:mera}
The Multiscale Entanglement Renormalization Ansatz builds on the TTN architecture by addressing one of its main limitations: the accumulation of short-range correlations during coarse-graining \citep{PhysRevLett.99.220405}.
To solve this, MERA incorporates \emph{disentanglers}: local unitary operators that filter out correlations between neighboring nodes without affecting their local dimensions.
By structurally alternating these disentanglers with isometries (which do alter the dimensions), the network prevents simulation collapse and achieves a much more expressive multiscale representation.
Consequently, MERA is particularly effective at capturing long-range dependencies and scale-dependent patterns in high-dimensional data, all while allowing for efficient contraction \citep{Reyes2021}.

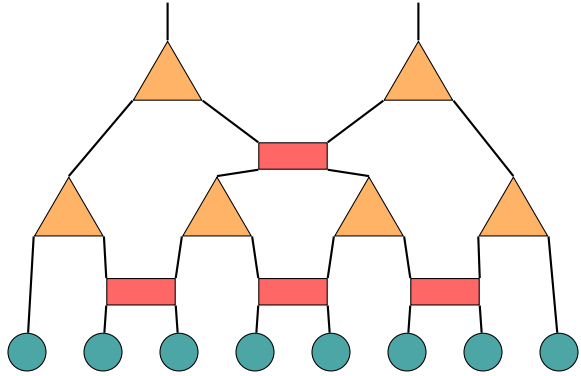
\begin{figure}[h]
    \centering
    \begin{tikzpicture}[
            unitary/.style={
    draw, 
    rectangle, 
    fill=red!60,
    text=white,
    minimum width=0.9cm, 
    minimum height=0.35cm,
    inner sep=0pt
    }
        ]

        \node[tnnode] (L1) at (0,0) {};
        \node[tnnode] (L2) at (1,0) {};
        \node[tnnode] (L3) at (2,0) {};
        \node[tnnode] (L4) at (3,0) {};
        \node[tnnode] (L5) at (4,0) {};
        \node[tnnode] (L6) at (5,0) {};
        \node[tnnode] (L7) at (6,0) {};
        \node[tnnode] (L8) at (7,0) {};

        \node[unitary] (U1_1) at (1.5, 0.8) {}; 
        \node[unitary] (U1_2) at (3.5, 0.8) {}; 
        \node[unitary] (U1_3) at (5.5, 0.8) {}; 
        
        \draw[bond] (L2) -- (U1_1.south west);
        \draw[bond] (L3) -- (U1_1.south east);
        
        \draw[bond] (L4) -- (U1_2.south west);
        \draw[bond] (L5) -- (U1_2.south east);
        
        \draw[bond] (L6) -- (U1_3.south west);
        \draw[bond] (L7) -- (U1_3.south east);

        \node[isometry] (V1_1) at (0.55, 1.8) {};  
        \node[isometry] (V1_2) at (2.5, 1.8) {};  
        \node[isometry] (V1_3) at (4.5, 1.8) {};  
        \node[isometry] (V1_4) at (6.4, 1.8) {};  
        
        \draw[bond] (L1) -- (V1_1.left corner);
        \draw[bond] (U1_1.north west) -- (V1_1.right corner);
        
        \draw[bond] (U1_1.north east) -- (V1_2.left corner);
        \draw[bond] (U1_2.north west) -- (V1_2.right corner);
        
        \draw[bond] (U1_2.north east) -- (V1_3.left corner);
        \draw[bond] (U1_3.north west) -- (V1_3.right corner);
        
        \draw[bond] (U1_3.north east) -- (V1_4.left corner);
        \draw[bond] (L8) -- (V1_4.right corner);

        \node[unitary] (U2_1) at (3.5, 2.6) {};
        
        \draw[bond] (V1_2.apex) -- (U2_1.south west);
        \draw[bond] (V1_3.apex) -- (U2_1.south east);

        \node[isometry] (V2_1) at (1.85, 3.6) {};
        \node[isometry] (V2_2) at (5.15, 3.6) {};
        
        \draw[bond] (V1_1.apex) -- (V2_1.left corner);
        \draw[bond] (U2_1.north west) -- (V2_1.right corner);
        
        \draw[bond] (U2_1.north east) -- (V2_2.left corner);
        \draw[bond] (V1_4.apex) -- (V2_2.right corner);

        \draw[phys] (V2_1.apex) -- ++(0, 0.5);
        \draw[phys] (V2_2.apex) -- ++(0, 0.5);
        
    \end{tikzpicture}
    \caption{MERA network. In contrast to TTNs, the disentanglers (blue rectangles) break short-range correlations so that the bond dimensions can accommodate long-range correlations.}
    \label{fig:mera_correct_legs}
\end{figure}

The multiscale nature of MERA makes it an exceptional feature extractor, particularly when relevant data is scattered across different resolutions \citep{Kong2021}.
The flexibility of these architectures has recently been expanded through branching variants, which open up multiple coarse-graining paths to enhance representational capacity \citep{Hou2024}.
Beyond classification, MERA decompositions have also been explored for structured detection problems; for instance, hyperspectral anomaly detection has been addressed by combining representations based on MERA with additional regularization priors \citep{Xiao_2024}.
In this context, unitary and hierarchical design principles, previously studied in tree-structured unitary TNs, provide useful guidance for stabilizing training and controlling information flow across scales \citep{Liu2019}.

\tocless\subsubsection{Tensor Decompositions}{}
So far, we have described architectures that are inspired by the study of quantum states of many particles.
However, in mathematics and numerical linear algebra, another set of tools for compressing high-dimensional data emerged in parallel: tensor decompositions.
Despite their different origins, both frameworks pursue the same goal: to represent and manipulate high-order tensors through low-rank structured factorizations.
This convergence has been highlighted in unifying perspectives that place tensor decompositions and TNs within a shared computational framework, especially in dimensionality reduction \citep{Cichocki2016, Cichocki2017}.
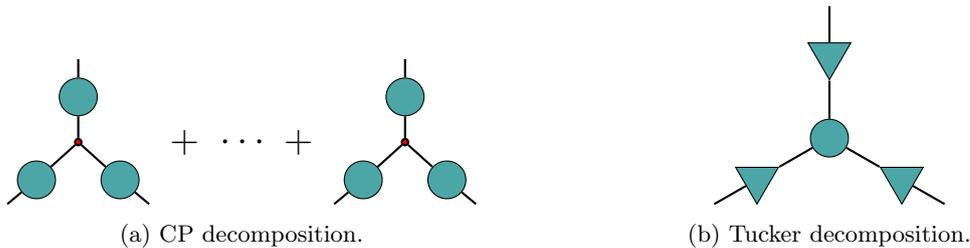
\begin{figure}[h]
    \null\hfill
    \begin{subfigure}[b]{0.45\textwidth}
        \centering
        \begin{tikzpicture}[scale=1.0]
            \node[tnnode] (A1) at (0,1.1)  {};
            \node[tnnode] (B1) at (-0.55,0) {};
            \node[tnnode] (C1) at (0.55,0) {};
            \node[draw,circle,thick,fill=red,inner sep=0.85pt] (R1) at (0,0.5) {};
            \draw[thick] (A1)--(R1);
            \draw[thick] (B1)--(R1);
            \draw[thick] (C1)--(R1);
            \draw[thick] (A1) -- ++(90:0.5);
            \draw[thick] (B1) -- ++(220:0.5);
            \draw[thick] (C1) -- ++(320:0.5);
            \node at (1.4,0.5) {\Large$+$};
            \node at (2.2,0.5) {\Large$\cdots$};
            \node at (2.9,0.5) {\Large$+$};
            \node[tnnode] (A2) at (4.3,1.1)  {};
            \node[tnnode] (B2) at (3.75,0) {};
            \node[tnnode] (C2) at (4.85,0) {};
            \node[draw,circle,thick,fill=red,inner sep=0.85pt] (R2) at (4.3,0.5) {};
            \draw[thick] (A2)--(R2);
            \draw[thick] (B2)--(R2);
            \draw[thick] (C2)--(R2);
            \draw[thick] (A2) -- ++(90:0.5);
            \draw[thick] (B2) -- ++(220:0.5);
            \draw[thick] (C2) -- ++(320:0.5);
        \end{tikzpicture}
        \caption{CP decomposition.}
        \label{fig:cp}
    \end{subfigure}
    \hfill
    \begin{subfigure}[b]{0.45\textwidth}
        \centering
        \begin{tikzpicture}[scale=1.0]
            \node[triangle, rotate=-90, minimum size=0.65cm] (A) at (90:1.1)  {};
            \node[triangle, rotate=-210, minimum size=0.65cm] (B) at (210:1.1) {};
            \node[triangle, rotate=-330, minimum size=0.65cm] (C) at (330:1.1) {};
            \node[tnnode] (G) at (0,0) {};
            \draw[thick] (A)--(G);
            \draw[thick] (B)--(G);
            \draw[thick] (C)--(G);
            \draw[thick] (A) -- ++(90:0.65);
            \draw[thick] (B) -- ++(210:0.65);
            \draw[thick] (C) -- ++(330:0.65);
        \end{tikzpicture}
        \caption{Tucker decomposition.}
        \label{fig:tucker}
    \end{subfigure}
    \hfill\null
    \caption{Graphical representations of common tensor decompositions.}
    \label{fig:tensor_decompositions}
\end{figure}
In modern data-driven applications, the most common tensor decompositions are the canonical polyadic (CP), Tucker, and tensor train (TT) decompositions.
In practice, these decompositions are used for low-rank compression and robust modeling, either by themselves or combining several ones \citep{Qiu2022}.
The CP decomposition (Fig.~\ref{fig:cp}), introduced in \cite{carroll1970analysis}, represents a tensor as the sum of rank-one components and is widely used due to its interpretability and parametric efficiency, including supervised learning formulations \citep{Haliassos2021}.
The Tucker decomposition (Fig.~\ref{fig:tucker}) generalizes matrix SVD to higher-order tensors, factorizing data into a compact central tensor and mode factor matrices \citep{tucker1966some}.
Aside from these, some of the architectures that we have discussed earlier have also been discovered in this context.
The Tensor Train (TT) decomposition was developed in \cite{Oseledets2011I}, and is algebraically equivalent to the MPS presented in Section~\ref{sec:mps} and depicted in Fig.~\ref{fig:mps_extended} \citep{Cichocki2016}.
Closely related, the hierarchical Tucker decomposition was originally introduced in \cite{hackbusch2009new} and organizes tensor modes into a tree structure and is equivalent to the TTNs discussed in Section~\ref{sec:ttn} and shown in Fig.~\ref{fig:ttn_8}.

In the ML literature, these decompositions have been adopted both as standalone models and as building blocks within TN-inspired architectures.
TT and hierarchical Tucker decompositions, in particular, enable the tensorization and compression of NNs, yielding compact models with reduced parameter counts and competitive performance \citep{novikov2015tensorizing}.
They also support streaming and online approximation settings, which are important for large-scale and dynamic data, as demonstrated by recent developments in incremental TT algorithms \citep{Kressner_2023}.
More generally, hierarchical tensor decompositions provide scalable multilevel representations for learning tasks \citep{Zniyed2020}, while advances in tensor–tensor algebra introduce alternative decomposition paradigms and optimality principles for multidimensional data representation \citep{Kilmer2021}.
Within the broader scope of tensor networks for ML, tensor decompositions thus constitute a mathematically grounded and complementary toolkit that both informs and benefits from developments in physics-inspired network architectures.

\section{Applications of tensor networks in machine learning}  \label{sec:TN_apps}
Now we proceed to analyze, from a broad perspective, the areas within ML where TNs are being actively used.
We have compiled the state-of-the-art bibliography available in the academic database OpenAlex, sorting the relevant articles by the domains of expertise identified during the analysis.
The details of the methodology followed can be found in Appendix \ref{sec:bib_analysis}, and a summary of the compiled database can be found in Fig.~\ref{fig:graph}.
The nodes in the graph denote authors (in pink) and topics (in green), and edges correspond to authorship relations.
An interactive version of this graph is also available, allowing the reader to explore the connections and to access the works in a dynamic manner \citep{valverde2026researchgraph}.
In the following, we proceed to analyze the different areas of application, showing concrete examples and critical assessments of the potential impact.

\subsection{Compression}\label{sec:compression}
Deep learning research constitutes a central focus within the study of TNs, which have been utilized to parameterize specific layers in deep neural networks, offering substantial advantages in computational efficiency and representational capacity.
A key driver of this interest is that deep learning is very intensive in terms of the memory necessary to store and evaluate the models, which motivates the use of TNs as a principled compression framework.
While other compression techniques such as pruning and quantization \citep{liang2021pruning} address this problem from complementary angles, TN-based methods are compatible with them and their combination can achieve higher compression rates (see, e.g., \cite{tomut2024compactifai}).

Thus, one of the most studied applications in this context is the replacement of large weight matrices and convolutional filters with low-rank approximations.
TNs have been used to compress dense layers \citep{novikov2015tensorizing}, convolutional filters in architectures such as ResNet-18 \citep{Yin_2021}, and recurrent networks \citep{Tjandra2017,Yang2017}, with reductions that lead to compression rates ranging from a factor of seven in \cite{novikov2015tensorizing} to a factor of forty in \cite{Tjandra2017}.
Beyond raw compression, TN approximations of linear layers have also been shown to alleviate overfitting \citep{Chen_2024}.
Connecting to modern architectures, these techniques have been extended to Transformers via Tucker and CP decompositions \citep{Ma2019}, and to embedding layers, thereby reducing the size of the dictionaries required in NLP (see Sec.~\ref{sec:deeplearning}).
In all these cases, parameter reduction is achieved while preserving competitive accuracy levels \citep{Gao2020,Zou2019,Jin_2020,Zhang2021}.

The structural affinity between TNs and high-dimensional data makes them a natural tool for data representation and processing at scale.
This representational efficiency extends to dataset compression: TNs have been applied to compress individual images \citep{latorre2005}, as well as image datasets and high dimensional data arrays \citep{Kilmer2021}, and to compress hyperspectral image data to improve their spectral resolution for downstream analysis \citep{Otgonbaatar_2023}, and to denoise and robustly represent tensor-structured data via hybrid Tucker-TT models \citep{Qiu2022}.
Beyond static datasets, TN-based methods have also been extended to distributed and streaming settings. In distributed settings, the data tensor is partitioned across multiple computing nodes, each handling a subset of the data, and the TT decomposition is computed in a distributed manner from these partial computations \citep{Wang2020}. In streaming settings, rather than reprocessing the entire dataset whenever new data arrives, the TT representation is updated incrementally to incorporate the newly observed data \citep{Liu2018, Kressner_2023}.
Their ability to operate in distributed and streaming settings, combined with the compression properties discussed above, positions TNs as a relevant tool for large-scale, continuous data environments.

The examples above establish TNs as a mature compression tool.
However, a critical limitation remains: tensorizing a pre-trained network is neither systematic nor robust.
It requires low-level knowledge of the architecture, careful selection of decomposition ranks, and typically demands fine-tuning after compression to recover accuracy.
While new approaches that do not need so much knowledge are being investigated \citep{pareja2025}, this sensitivity makes deployment non-trivial and motivates the search for more principled approaches to TN-based model design.

\subsection{Explainability and privacy}
\label{sec:exp}
TNs possess structural properties that make them particularly relevant in contexts where transparency and privacy are critical requirements; two concerns that are often in tension in classical ML architectures.

From an interpretability standpoint, unlike traditional black-box models, hierarchical TNs such as TTNs or MERA are designed to learn and extract relevant features in a structured and traceable manner \citep{Stoudenmire2018}.
This property makes them a suitable alternative for developing ML models that, in addition to being computationally efficient, offer a high degree of interpretability, as reviewed in \citep{Ran2023}.
In particular, MPS-based architectures are being explored in Natural Language Processing tasks to allow for greater traceability in how models represent and process text sequences \citep{Tangpanitanon2022}.
Moreover, TN models can use tools such as von Neumann entropy and quantum mutual information to analyze feature importance, making them a compelling option for explainable AI in critical environments \citep{aizpurua2025tensor}.
Along similar lines, recent theoretical work shows that TNs also admit efficient feature-attribution analyses: SHAP values, which quantify the contribution of each input feature to a prediction and are NP-hard to compute for general NNs, can be computed exactly for TN architectures, and in polylogarithmic time for TT structures \citep{marzouk2025shap}.

From a privacy standpoint, current ML models are vulnerable to adversarial data-extraction attacks, where an attacker can retrieve sensitive training data.
The standard procedure to avoid this issue is to train the models using differential privacy techniques \citep{dwork2006differential,dwork2014algorithmic}.
These typically involve adding carefully crafted noise, that impacts trainability and performance of the final model.
In contrast, the gauge freedom of TNs (see Sec.~\ref{sec:gauge}) enables a form of model obfuscation that is intrinsic to the architecture.
This can be achieved by training native TN models \citep{Pozas-Kerstjens2024A} or by tensorizing pre-trained neural networks \citep{pareja2025}.
Importantly, this mathematical restructuring can enhance privacy while preserving, and in some cases improving, interpretability, thereby mitigating the trade-off often observed in classical architectures.

These properties are particularly relevant in healthcare, where both interpretability and data privacy are regulatory requirements.
In this context, explainable TN-based ensemble learning has been applied to the dynamic prediction of Alzheimer's disease from brain structure variation \citep{zhang2022explainable}, and tensorized GANs with high-order pooling have been used for Alzheimer's disease assessment \citep{Yu_2022}.

In summary, TNs are among the few architectures that simultaneously combine efficiency, interpretability, and privacy.
They are therefore strong candidates in fields where transparency is not optional but a requirement.
However, the literature still lacks standard metrics and protocols for evaluating the degree of interpretability offered by these architectures, and empirical validation in complex real-world models remains limited.

\subsection{Integrations in deep learning}\label{sec:deeplearning}
Beyond the applications in model and data compression explained in Section~\ref{sec:compression}, TNs have been integrated into deep learning architectures across multiple domains, such as computer vision and natural language processing, where their ability to preserve high-dimensional structure and capture multilinear dependencies offers advantages over standard neural network approaches.
This integration includes hybrid architectures that combine TNs and NNs in a common framework, exploiting the natural correspondence between TNs and probabilistic graphical models \citep{glasser2020probabilistic}.
Furthermore, CNNs and RNNs have been shown to possess equivalent TNs representations that exhibit volume-law entanglement scaling \citep{Levine2019}, establishing a theoretical bridge between deep learning expressivity and the entanglement structure of tensor networks.

In computer vision, one of the most impactful applications is the tensorization of convolution operators in CNNs, which significantly reduces the number of model parameters without compromising representational capacity. This strategy has been applied to tasks such as multifunctional biometric recognition \citep{Jin2020} and the classification of tiny objects via quantum-inspired TN architectures based on MERA \citep{Kong2021}. For higher-order sequential visual data such as videos, where parameter counts explode with dimensionality, TN-based architectures provide a natural and efficient solution, as demonstrated by TT-based hierarchical RNNs for video summarization \citep{Yang2017,Zhao2020} and by quantized TT networks for 3D object and video recognition \citep{Lee2021}. In more complex settings such as hyperspectral image processing, where data is naturally represented as high-dimensional tensors spanning tens or hundreds of electromagnetic bands, TNs provide an appropriate architecture for modeling and compression. Applications include spatial anomaly detection using advanced architectures such as MERA \citep{Xiao_2024}, temporal change analysis \citep{Zhou_2020}, and multi-resolution image fusion \citep{Yang2023}. Furthermore, TN-based circuits have been extended to quantum computing for image classification \citep{Guala2023}. These techniques have also proven valuable in high-stakes biomedical applications: \citet{Yu_2022} combine TN-based GANs with high-order pooling to improve early detection of Alzheimer's disease from magnetic resonance images, demonstrating that structural integration of TNs in complex architectures can yield both efficiency and clinical relevance.

In NLP, the compression of embedding layers represents one of the most concrete successes of TNs. A TT-embedding configuration achieved a 441× reduction in parameters on an Internet Movie Database (IMDB) sentiment classification task while simultaneously increasing accuracy by 1\% compared to a standard embedding layer \citep{Hrinchuk2020O}.
Similarly, in the Workshop on Machine Translation (WMT) 2014 English-German machine translation task, a TT-embedding with a 15× compression factor resulted in only a 0.3 drop in BiLingual Evaluation Understudy (BLEU) score \citep{Ma2019}.
More recently, \cite{yang2024loretta} introduced a TT-based adaptation of LoRA \citep{hu2022lora}, a parameter-efficient fine-tuning technique that replaces full weight updates with low-rank matrix increments, achieving competitive downstream performance with orders of magnitude fewer trainable weights.

Despite this wide applicability, important limitations remain.
Benchmarking against state-of-the-art vision and language models is still insufficient, making it difficult to draw definitive conclusions about the competitiveness of TN-based architectures.
As in compression, the selection of appropriate TN formats and ranks for a given task requires significant domain expertise, and more systematic approaches to architecture design remain an open research direction.

\subsection{Supervised learning}
The largest number of applications of TN methods in ML, according to our analysis, have to do with supervised learning tasks.
Beyond those already discussed in the previous sections, which were based on tensorized versions of conventional ML architectures, here we discuss further concrete uses for classification and regression using TNs as the learning models.

A foundational contribution is \citet{Stoudenmire2016}, which demonstrated how MPS can parameterize a supervised classifier via a local quantum-inspired feature map, optimized with a DMRG-style sweeping algorithm.
A closely related approach, developed independently from a Tensor Decomposition angle, is that of Exponential Machines \citep{novikov2018exponential}, which use the TT format to 
implicitly represent all feature interactions up to arbitrarily high order, trained via stochastic Riemannian optimization.
The connection between TNs and probabilistic graphical models has also been exploited for supervised learning, motivating generalized TN architectures where information can be copied and reused across the network, overcoming the limitations of regular TNs in higher dimensions \citep{glasser2020probabilistic}.

In image classification, several studies applied TNs on widely used datasets such as MNIST and Fashion-MNIST \citep{Chen2017,Wall2021,sun2020generative}.
A prominent example is the use of TTN where most layers are trained in an unsupervised manner via coarse-graining and only the top layer is optimized for classification, achieving an accuracy of 89.9\% on Fashion-MNIST \citep{Stoudenmire2018}.
Two-dimensional hierarchical TNs with unitary tensors, optimized via a MERA-derived algorithm, have also been trained for image recognition, overcoming the scalability limitations of one-dimensional architectures \citep{Liu2019}.
Supervised learning with PEPS has also been explored, motivated by the two-dimensional structure of images \citep{Cheng2021}; the model matches the performance of feed-forward NNs with significantly fewer parameters and outperforms tree-like TNs on the same benchmarks.
Video classification has also been addressed using tensor-train RNNs trained in an end-to-end manner \citep{Yang2017}. 

Beyond visual data, TNs have been applied to high-energy physics data classification \citep{Felser2021}, and to multi-scale classification and regression tasks \citep{Reyes2021}, where the ability of TNs to handle complex multidimensional data improves efficiency in extracting temporal and frequency features.
In the case of regression, TNs have shown promising results in time series analysis, with applications ranging from financial forecasting \citep{yao2021} to urban traffic flow prediction \citep{wu2021tensor}, exploiting the ability of TT-RNNs to model long-range dependencies at a fraction of the parametric cost of standard RNNs.

However, evaluations in the literature are still largely based on small datasets such as MNIST and Fashion-MNIST, while evaluations on modern large-scale benchmarks remain scarce; until this gap is addressed, the practical potential of TN-based supervised models remains unclear.

\subsection{Unsupervised learning}
Although unsupervised learning has received less attention than supervised learning, there are also relevant contributions in this context, that range from generative modeling to anomaly detection or feature extraction.

TNs have been applied to generative modeling through Born machines \citep{han2018}, which represent probability distributions by using the quantum-state representation of tensor networks.
Namely, the parameters of the network encode not the probabilities themselves, but a wavefunction from which probabilities are obtained by squaring the corresponding coefficients \cite{nielsen2010quantum}.
One important advantage of these architectures is that using algorithms such as two-site DMRG the bond dimension can be adjusted dynamically during training, allowing the model's representational capacity to grow adaptively with the data.
This is in stark contrast to classical generative models such as Boltzmann machines, where the architecture is fixed prior to training.
While initially developed within the MPS formalism in \cite{han2018}, Born machines were quickly extended to trees \citep{Cheng2019}, as well as several other architectures simulating quantum circuits \citep{Huggins_2019}.
A rigorous analysis of the expressive power of these probabilistic TN-based models has further shown that different TN formats exhibit provable separations in representational capacity \citep{Glasser2019I}.

More recently, this framework has been extended to continuous data \citep{Meiburg2025} and to diffusion-based processes \citep{Causer2025}, broadening the scope of TN-based generation beyond discrete distributions.
Furthermore, TTNs have been applied to the generation of molecular datasets \citep{Cheng2019,Moussa_2023}, and models based on MPS have demonstrated the ability to perform classification and generation simultaneously \citep{mossi2025}.

Another notable application within unsupervised learning is anomaly detection \citep{wang2020anomaly}, which is present in different fields.
Indeed, it has been used in problems that range from the identification of events in proton-proton collisions in particle accelerators \citep{puljak2025tensor}, to detecting spatial and spectral changes in hyperspectral images \citep{Zhou_2020,Xiao_2024}, and to unsupervised compressed sensing tasks where TNs learn sparse signal representations without labeled data \citep{ran2020tensor}. Image fusion of hyperspectral and multispectral data has also been addressed with deep unsupervised TN architectures, exploiting the natural tensor structure of spectral data \citep{Yang2023}.

Similarly, MPS-based models can learn joint data distributions in an unsupervised manner by grouping information according to its statistical structure.
Unlike classical clustering methods such as $k$-means or Gaussian mixture models, the bond dimension of the MPS acts as a natural control over the complexity, capturing only the correlations present in the data without assuming a fixed parametric form for the distribution.
This results in more interpretable representations \citep{Shi2022,Liu_2023}.

Furthermore, scalable TN frameworks such as CP decompositions and multi-scale hierarchical TNs have proven effective in extracting latent features from heterogeneous high-dimensional data in an unsupervised manner \citep{Stoudenmire2018,Liu2019}, offering a natural multilinear alternative to classical dimensionality reduction methods such as PCA.

In summary, unsupervised TN methods have shown genuine theoretical strengths in generative modeling and in anomaly detection, where learning data distributions without labels is a natural fit for identifying rare events. Feature extraction with multi-scale TNs also offers a structured alternative to classical dimensionality reduction. That said, empirical validation remains limited: most results are confined to small or domain-specific datasets, systematic comparisons with modern deep generative models are scarce, and applications to clustering and semi-supervised learning remain largely underdeveloped despite theoretical motivation.

\subsection{Quantum machine learning}

The remarkable ability of TNs to approximate states of quantum systems of many particles has led to them being the state of the art in the classical simulation of quantum computers \citep{feng2022,liao2023kicked,tindall2024,zhao2024,oh2024,begusic2024}.
As a consequence, TNs are now used to simulate quantum machine learning algorithms, especially those based on parameterized quantum circuits (PQCs).

These methods have given rise to a different type of ``quantum-inspired'' algorithms, where the algorithm itself is fully quantum but the execution is classical via TNs.
Unlike conventional quantum machine learning models, which are restricted to unitary transformations, these TN-based algorithms admit not-unitary factorizations that grant them with extended capabilities.
For example, locally purified states have been proven to be strictly more expressive than their unitary components, with unbounded separations in resource requirements \citep{Glasser2019I}.
By mapping PQC architectures to tree or MPS-based TNs, one obtains qubit-efficient circuits whose resource requirements scale only logarithmically with input size, while also showing resilience to noise \citep{Huggins_2019}.
Applications beyond theoretical can also be found, using TN-based PQCs to solve real-world tasks such as image classification \citep{Guala2023} and hybrid classical-quantum classifiers where part of the processing is handled classically via TNs and the other part on quantum hardware \citep{Chen_2021,Hou2024}.
Moreover, physical insights such as decoherence and postselection can be exploited to develop better models at lower costs \citep{Liao2023}.

On the flip side, TNs also contribute to the development of ML algorithms that will run on quantum computers.
For example, TNs enable efficient encoding of classical data onto quantum computers \citep{ran2020encoding,Rieser2023,kiwit2025typical}.
In its most extreme version, TN models are trained classically and then deployed on quantum hardware \citep{wall2021generative}.
Still, one of the main advances has been the pretraining of PQCs via TNs: by initializing circuit parameters using MPS-based learning, convergence in supervised tasks is substantially accelerated \citep{Dborin_2022}, and initialization issues such as barren plateaus are mitigated \citep{Rudolph_2023}.
The barren plateau problem is also directly addressed by isometric TN architectures, where gradients have been proved to avoid exponential vanishing even in large-scale setups \citep{barthel2025absence}.

Currently, TNs are the strongest link between classical ML and quantum computing, both theoretically and practically.
They facilitate the encoding of classical data onto quantum hardware, provide benchmarks to evaluate genuine quantum advantage \citep{jozsa2006simulation}, address fundamental training challenges, and serve as a theoretical lens for understanding when a quantum model can be efficiently replaced by a classical one \citep{shin2024dequantizing}.
The bidirectional nature of this relationship is one of its key strengths: advances in TNs inform quantum circuit design and vice-versa.

There are also applications of TN-ML in more physical areas, such as quantum simulation, as well as work that combine both techniques to study theoretical foundations such as entanglement or quantum mutual information.
However, as this falls outside the scope of this work, we have compiled several relevant articles on these topics, which can be found in the interactive graph under the subtopics ``Quantum Simulation'' (those of a more physical nature) and ``Quantum Algorithms and Theory'' (those of a more theoretical nature) \citep{valverde2026researchgraph}.

\section{Software solutions}\label{sec:soft}
A crucial ingredient that is necessary to obtain all the results discussed above is numerical implementations of TNs and tensor decompositions.
Strictly speaking, no specific software packages are necessary: given their mathematical transparency, TNs can be implemented directly using native tensor operations.
For instance, in Python, libraries like NumPy \citep{harris2020array} or PyTorch \citep{ansel2024} provide the \texttt{einsum} operation for contractions, as well as operations such as \texttt{reshape}, \texttt{permute}, \texttt{transpose}, \texttt{svd}, \texttt{qr}, and \texttt{tensordot}, that are sufficient to build and manipulate most TN architectures.
However, doing so requires familiarity with index ordering conventions and consistent notation throughout the network.
Particular attention must be paid to the contraction path (the order in which tensors are contracted), since suboptimal choices can introduce exponential memory overheads.
Despite the fact that finding the optimal contraction path is an NP-hard problem \citep{lam1997,arad2010}, tools like \texttt{opt\_einsum} \citep{Smith2018} can help in finding good contraction paths by using various heuristic methods.
These are the subtleties that are addressed by the libraries that we discuss below.

Early solutions are rooted in many-body physics, such as ITensor \citep{fishman2022itensor} (in Julia) and TeNPy \citep{tenpy} (in Python).
These libraries laid the algorithmic foundations of TN optimization through methods such as DMRG and TEBD.
Google's TensorNetwork \citep{roberts2019tensornetwork} later brought these ideas closer to ML, although the library is now obsolete.
The two most mature solutions today are TensorLy \citep{kossaifi2019tensorly}, geared towards tensor decomposition with support for multiple Python backends (NumPy, PyTorch, TensorFlow), and TensorKrowch \citep{monturiol2024tensorkrowch}, focused on TN-based architectures and native TN models.
The latter is written on top of PyTorch, thereby allowing for interaction with the rest of the library and enabling, for instance, the development and training of hybrid architectures that combine tensor networks and neural network layers.
Both include detailed tutorials and cover the differentiable programming required by these architectures. More specialized options include TedNet \citep{pan2022tednet}, which provides ready-to-use tensorized layer implementations directly in PyTorch, and tntorch \citep{usvyatsov2022tntorch}, which features full autodiff support for CP, Tucker, and TT formats, as well as additional tools for tasks such as elementwise tensor arithmetics or sensitivity analysis.

In the context of quantum machine learning, there are additional options.
Quimb \citep{Gray2018} allows explicit manipulation of TN architectures with automatic differentiation, including quantum circuit simulation and support for high-dimensional quantum systems.
In the quantum computing ecosystem, PennyLane \citep{bergholm2018} integrates TNs as a simulation backend and allows differentiation through the quantum circuits that are simulated, with native integration in PyTorch and JAX for training hybrid TN-quantum models.
Qiskit \citep{javadi2024} also allows MPS to be used as a simulation backend, although without direct access to the underlying TN.
Finally, it is worth mentioning cuTensorNet (\citeauthor{cutensornet}), which, with the aim of accelerating both quantum simulation and classical inference, offers efficient contraction of TNs on GPUs.

\section{Discussion}  \label{sec:discussion}
The architectures and applications reviewed in this paper paint a consistent picture: Tensor networks offer provable advantages for machine learning in terms of efficiency, interpretability, and privacy.
However, despite an increasing number of proof-of-concept demonstrations, they are largely absent from the state of the art in AI and from critical sectors where these properties are necessary (e.g., banking, healthcare, etc.). The central argument of this discussion is clear: this gap is not due to a fundamental theoretical limitation, but a consequence of the current structure of the field.

To understand this limitation, we must first clarify where TNs have a proven advantage.
The main one is what they were designed for: efficiently handling spaces with high local structure or low correlation between features.
However, despite this geometric intuition, TNs have proven to be a powerful technique for compressing deep learning models such as LLMs: although their weights are dense tensors, training tends to converge to low-rank solutions, revealing exploitable parametric redundancy.
This suggests that neural networks are overparameterized by design, and that TNs are a more natural tool for learning while avoiding this redundancy.
However, most demonstrations available are small-scale, such as in MNIST or toy datasets.
We attribute this, at least in part, to the historical development of the field, which has been led mainly by physicists and mathematicians. As a result, progress has often prioritized formal understanding and algorithmic elegance over software usability, industrial performance, and validation on real-world use cases.

This gap between theory and practice is especially striking when considering the sectors that most need what TNs offer by design.
Banking, healthcare, and critical infrastructure (to mention a few) are domains where privacy and interpretability are not optional, yet TNs are practically absent from them.
The paradox intensifies when one realizes that these properties are not techniques added after the fact, but inherent consequences of the mathematical structure of the formalism: gauge freedom allows control over the internal representation, canonical forms make information locally accessible, and tools inherited from quantum information theory offer precise metrics on what information the model captures, how it is processed, and how different features interact with each other.
Why haven't these advantages translated into adoption? Not because of theoretical limitations: rather, it shows that the field still needs to incorporate more feedback and expertise from communities closer to real-world deployment, such as applied machine learning, software engineering, and hardware design.

The same regime that explains the advantages of TNs as promising ML architectures (low rank and low correlation between features) is precisely what characterizes quantum computing in the current era of noise and small-scale devices.
Noise limits the achievable entanglement, so TNs are the natural tool for studying and optimizing models in this regime.
Thus, TNs allow the creation of a sandbox where quantum algorithms can be simulated in a controlled manner, pushing models to their theoretical limits until high-quality quantum hardware is available.
This makes TNs an autonomous research technique in quantum machine learning, not a bridge.
Furthermore, the TN formalism allows the development of quantum-inspired algorithms already useful in classical hardware, such as Born machines, which exploit the structure of quantum amplitudes to model probability distributions with formal guarantees that classical models cannot offer by design.
Regardless of how quantum computing evolves, TNs will remain relevant for quantum advantage benchmarking, efficient classical simulation, circuit pretraining, and classical data encoding.

Despite the notable advances, the field of TNs is still in active development.
While most of the improvements have come from physicists and mathematicians interested in theory (and we expect that this will be also the case in the future), the new wave of advances must be more focused on practice.
A primary one concerns compression: local SVD guarantees that the dimensions with higher variance are retained in TN approximations.
However, in ML the individual variance of a matrix is not necessarily the relevant criterion for the downstream task.
The usual approach (post-compression fine-tuning) is a workaround, not a solution. This opens up an unexplored avenue: task-, architecture-, and data-informed surgical compression, acting directly on the internal structure of the model with criteria richer than the matrix variance.
Another one with high potential impact concerns the structure of deep architectures: in deep learning models, nonlinear activation functions are applied point-wise.
When these models contain TN layers (because of design or as a consequence of tensorization), this forces the reconstruction of the dense tensor before applying them.
Finding a tensor representation of the nonlinearities is likely the most important open technical problem for accelerating the adoption of these architectures.
On the hardware side, GPUs and TPUs are optimized for dense matrix multiplication, especially in large and regular workloads, rather than for the many small structured contractions that often arise in TNs models. As a result, the theoretical computational advantage of tensorized models (which substitute large matrix multiplications with multiple smaller ones) does not necessarily translate into a reduction in inference time unless the bond dimension is very low or these contractions can be efficiently batched and parallelized.
Realizing that advantage requires development in specialized architectures such as FPGAs or ASICs, which opens a line of research closer to high-performance engineering than to theory.
Finally, regarding interpretability, guarantees via gauge freedom apply to TN-native architectures, and not necessarily to a posteriori tensorizations of neural networks.
This distinction is key, and currently one that the literature frequently overlooks but that nevertheless has real-world consequences in regulated sectors.

This situation is not new.
Tensor networks find themselves today where neural networks were before GPUs and backpropagation: solid theoretical foundations, but lacking the infrastructure that makes the advantage accessible.
The difference between consolidating as a general-purpose tool or remaining a niche alternative will be determined by the ecosystem.
For this leap to occur, specific conditions are needed: libraries with layers of abstraction that allow the design of models without deep mathematical expertise, specialized hardware that materializes the theoretical computational advantage, and systematic collaboration between physicists, mathematicians, and software and hardware engineers.
Added to this is the need to communicate their applications beyond academia.
In this sense, the increasing energy cost of language models is the most accessible argument: presenting TNs as a way to make AI more efficient opens the conversation to sectors that would not otherwise have it.
If these conditions are met, TNs may be able to reach their full potential as a practical ML framework.
If not, they will remain powerful but marginal.

\section*{Acknowledgements}
This work is supported by the Basque Government through the Elkartek project KUBIBIT- kuantikaren berrikuntzarako ibilbide teknologikoak (ELKARTEK25/79), the Spanish Government grant PID2023-146758NB-I00 funded by MICIU/AEI/10.13039/501100011033, and the Swiss National Science Foundation (grant number 224561).

\bibliography{references} 

\appendix
\section{Methods} \label{sec:bib_analysis}
In order to perform the bibliographic analysis, we compiled the state-of-the-art bibliography using OpenAlex (\citeauthor{OpenAlex}), an academic database that allows to perform advanced searches with specific filters.
This enables us to focus on academic articles on TNs and their variants (such as MPS, PEPS, and TT) while covering a broad range of ML subfields, including Supervised learning, NNs, DL, and more.

The OpenAlex database provides a comprehensive set of 184 features, which enabled us to apply diverse filtering criteria such as publication source, citation count, and relevance score.
Among these attributes are unique identifiers (ID, DOI), publication metadata (title, date), and bibliometric indicators like the Field-Weighted Citation Impact (FWCI) and citation count.
Additionally, OpenAlex offers details on open-access status, licensing, authors’ institutional affiliations, and thematic categorization.
This rich metadata has been crucial in ensuring that our dataset consists solely of high-quality journal or conference publications, guaranteeing its validity and relevance.

\begin{figure*}[t]
    \centering
    \includegraphics[width=\textwidth, height=\textheight, keepaspectratio,clip,trim={30cm 50cm 30cm 50cm}]{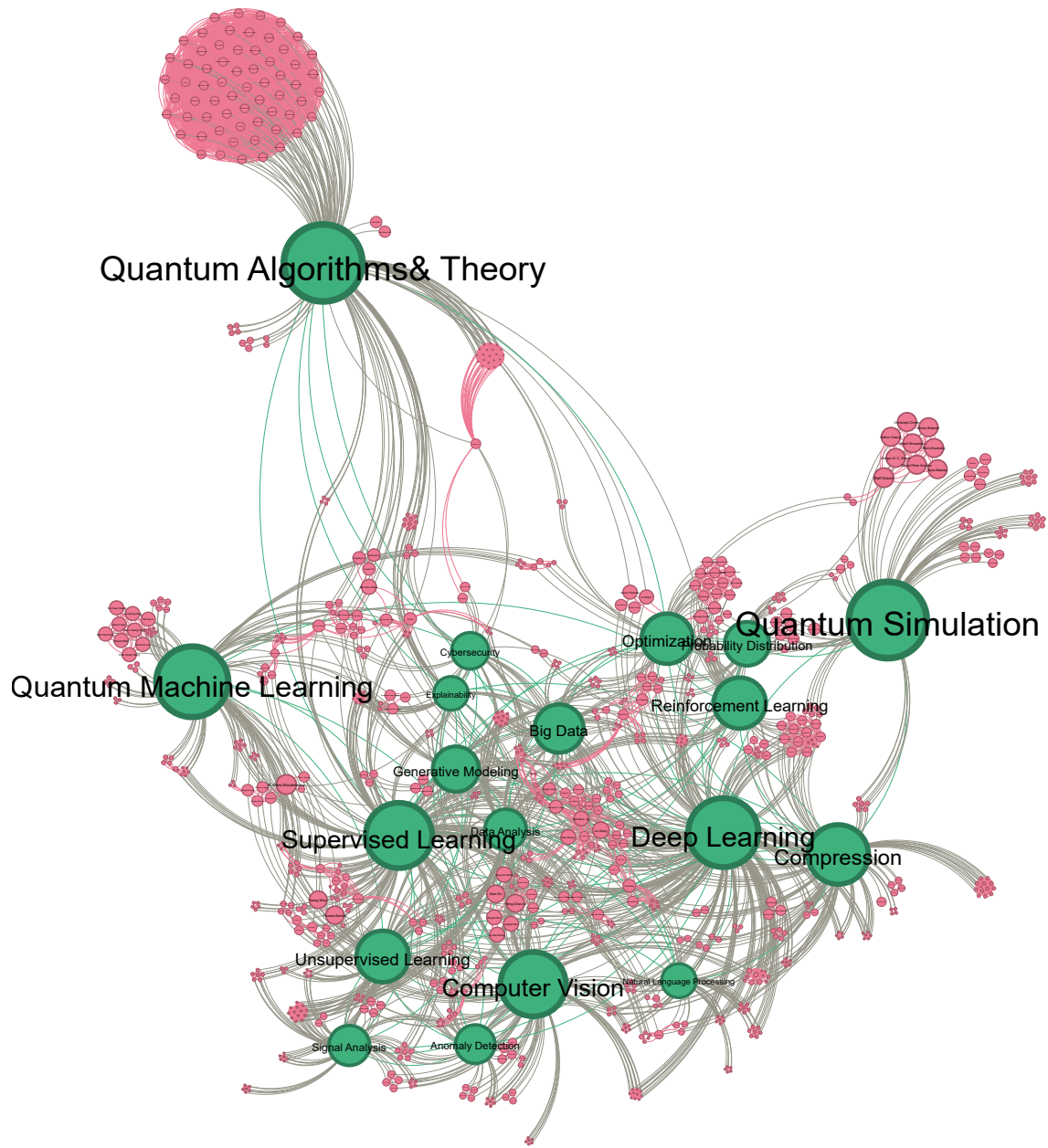}
    \caption{Topic-author graph for the works analyzed. Green nodes correspond to topics, while pink nodes denote individual authors. Two nodes are connected if an author has published a work in said topic. An interactive version of this graph, that allows for accessing the relevant manuscripts, can be found in \cite{valverde2026researchgraph}.}
    \label{fig:graph}
\end{figure*}

To ensure the rigor of our dataset, we applied detailed filters that initially yielded $1\,071$ articles.
Careful query design was crucial, as TN-related terms can appear in different formats (e.g., ``Tensor Network'' vs. ``Tensor-network'').
The complete query that leads to our dataset can be found in in \citep{OpenAlex2025}.
To this dataset we applied a series of filters that eliminated duplicates, irrelevant publications, and those not indexed in high-quality conferences or journals.
Our final dataset contains a total of 181 articles.
Using this curated database, we generated Fig.~\ref{fig:graph}, which illustrates the relationship between the subtopics of the filtered papers, weighted by their relevance, and most relevant authors in each subtopic.
As a result of the analysis, we do a clustering by ML-related topics, finding that TNs and tensor decompositions have been applied in a significant amount of areas, including supervised learning, quantum machine learning, deep learning, unsupervised learning, big data, computer vision, generative modeling, and natural language processing.
An interactive version of the graph can be found in \citep{valverde2026researchgraph}.

\end{document}